\documentclass[letterpaper,journal]{IEEEtran}

\usepackage{amsmath,amsfonts,amssymb}
\usepackage{array}
\usepackage{graphicx}
\usepackage{booktabs}
\usepackage{multirow}
\usepackage{makecell}
\usepackage{algorithm}
\usepackage{algorithmic}
\usepackage[table]{xcolor}
\usepackage[hyphens]{url}
\usepackage{cite}
\usepackage{stfloats}
\usepackage[hidelinks]{hyperref}

\newcommand{\cmark}{\ensuremath{\checkmark}}
\definecolor{urlblue}{RGB}{9,105,218}
\definecolor{refrose}{RGB}{199,21,133}
\hypersetup{
    pdfkeywords={Autonomous driving, contrastive learning, methods for safety, world-action models},
    colorlinks=true,       
    urlcolor=urlblue,       
    linkcolor=refrose,      
    citecolor=urlblue    
}

\title{Reinforcing End-to-End Autonomous Driving Safety via Negative-Aware Planning in a Unified World-Action Model}

\author{Zihao Zhang$^{1,2}$, Xinyu Zhang$^{1}$, Bing Zeng$^{1}$, Xingyu Chen$^{3}$, Zhenning Li$^{4}$, Bin Yu$^{5}$, Ziyuan Pu$^{1 \ddagger}$ \\[1.0em]
$^{1}$Southeast University\quad $^{2}$Shenzhen Loop Area Institute\quad $^{3}$Guangxi University \\
$^{4}$University of Macau\quad $^{5}$Beihang University \\
\thanks{$^{\ddagger}$Correspondence Author.}
}

\begin{document}

\maketitle

\begin{abstract}
    World-action models (WAMs) jointly model future traffic states and driving actions for end-to-end autonomous driving in intelligent transportation systems. However, their action learning remains dominated by expert-trajectory imitation, leaving unsafe alternatives close to expert trajectories insufficiently constrained. To improve WAM robustness in safety-critical scenarios, we propose \textbf{NegWAM}, a safety post-training framework for unified WAMs that learns explicit safety boundaries from near-expert failure trajectories. NegWAM constructs negative trajectories that are geometrically close to the expert trajectory yet carry safety risks, uses them as repulsive anchors to steer predictions away from dangerous behavior, and encodes them as scene-conditioned rider tokens to learn scene--trajectory compatibility. During training, NegWAM jointly optimizes the flow-matching and near-expert negative objectives, while the auxiliary rider branch is discarded at inference, incurring no additional overhead. On NAVSIM v2, NegWAM achieves 40.4 EPDMS on \textit{navhard}, a 4.1\,\% relative improvement over the strongest compared scorer-free WAM baseline, and achieves 93.3 EPDMS on \textit{navtest}. These results demonstrate that NegWAM achieves strong performance in safety-critical scenarios while preserving routine-driving performance. Zero-shot closed-loop evaluation on HUGSIM further demonstrates cross-benchmark generalization. The code is available at \url{https://github.com/Zizizi-hao/NegWAM}.
\end{abstract}


\section{Introduction}
\label{sec:intro}

End-to-end autonomous driving systems typically learn driving policies from large-scale expert demonstrations and perform well in routine scenarios~\cite{hu2023uniad, chen2024end-to-end}. However, hazardous behaviors and failure cases are scarce in real-world driving data, making it difficult for imitation learning to acquire explicit safety boundaries~\cite{karkus2025beyond, yang2026posttraining}. World-action models (WAMs) introduce stronger priors over scene dynamics into end-to-end autonomous driving by predicting future world states and leveraging future representations to guide planning~\cite{xia2026drivelaw, xu2026eponav2, shi2026drivewam}. Yet their action branches are still trained primarily through flow matching on expert trajectories. This supervision tells the model what to imitate, but not which near-expert alternatives should be rejected as unsafe. Consequently, stronger world modeling alone does not establish explicit safety boundaries, motivating safety-oriented post-training for WAMs.

\begin{figure}[!tbp]
\centering
\includegraphics[width=\linewidth]
{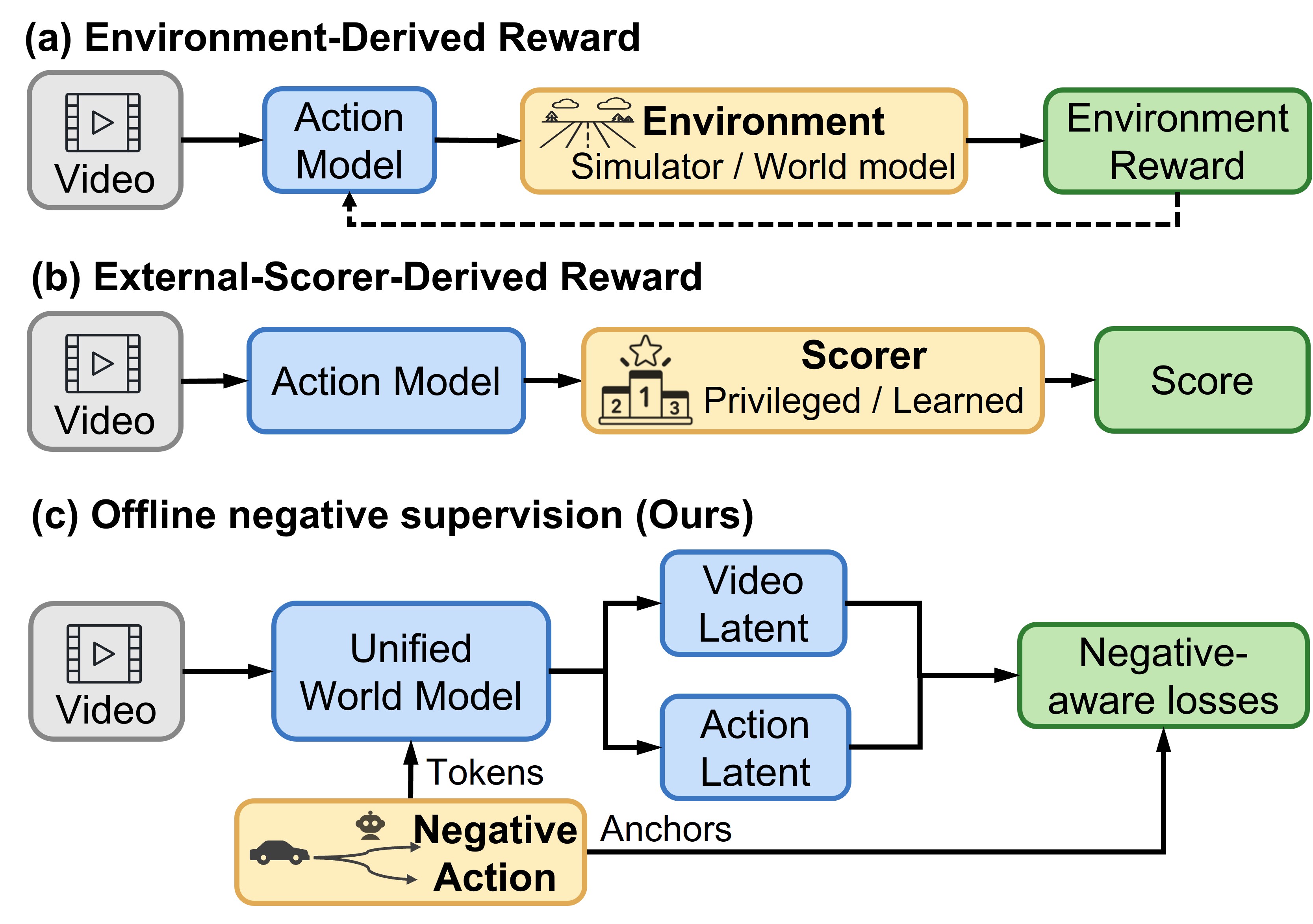}
\caption{
Sources of supervision for post-training in end-to-end autonomous driving.
(a) Environment-derived rewards from closed-loop interaction with a simulator or world model.
(b) Trajectory scores from an external rule-based or learned scorer.
(c) NegWAM uses offline-mined negative trajectories as repulsive anchors and scene-conditioned rider tokens to introduce negative-aware losses into a unified world--action model.}
\label{fig:fig1_teaser}
\end{figure}

Post-training offers a natural avenue beyond these limitations, but a central challenge is constructing learning signals that faithfully encode desired driving behavior~\cite{rafailov2023dpo, yang2026posttraining}. Existing methods primarily rely on two sources of supervision. One line of work (Figure~\ref{fig:fig1_teaser}a) explores policies in closed-loop environments built with a simulator~\cite{dosovitskiy2017carla, kerbl20233dgs} or a world model~\cite{yang2026resim, li2026world}, but is constrained by the sim-to-real gap, environment construction costs, and reconstruction artifacts~\cite{gao2026rad, li2026world}. Another line (Figure~\ref{fig:fig1_teaser}b) scores candidate trajectories using rule-based evaluators or learned reward models, but the quality of its supervision is limited by handcrafted rules, privileged information, and evaluator bias~\cite{yang2026dreamerad, yang2026worldrft, xu2026toad}. These limitations underscore the need for additional sources of effective supervision for autonomous driving post-training.

Unsafe driving behaviors provide critical cues for delineating safety boundaries and are often easier to characterize than the full range of desirable driving behavior~\cite{wang2026beyond, gao2026learning}. Building on this insight, we propose \textbf{NegWAM},  a negative-aware safety post-training method that incorporates offline negative supervision into a unified WAM (Figure~\ref{fig:fig1_teaser}c). Specifically, near-expert negatives serve as repulsive anchors that mark locally unsafe regions in trajectory space and as scene-conditioned rider tokens that enable shared attention layers to learn scene--trajectory compatibility. Together with the original imitation objective, these mechanisms enable the WAM to learn not only how experts drive but also which behaviors should be avoided in the current scene.

Unlike prior hard-negative learning that primarily regularizes planner outputs~\cite{wang2026beyond}, NegWAM post-trains both the single-rollout action flow and the shared world--action representation of a unified WAM. It mines near-expert trajectories that fail offline safety evaluation and jointly optimizes the original flow-matching objective with two hard-negative objectives. The evaluator is used only for offline mining, while the rider branch and compatibility head are removed at inference, adding no inference-time modules. On NAVSIM v2~\cite{dauner2024navsim,cao2025navsimv2}, NegWAM achieves 40.4 EPDMS on \textit{navhard}, a 4.1\,\% relative improvement over the strongest compared scorer-free WAM baseline, and achieves 93.3 EPDMS on \textit{navtest}, surpassing the strongest baseline score of 92.9 by 0.4 points and demonstrating stronger challenging-scenario performance without compromising routine driving. Zero-shot closed-loop evaluation on HUGSIM~\cite{zhou2025hugsim} further supports cross-benchmark transfer, with NegWAM achieving a higher HD-Score than DriveWAM (30.6 vs.\ 27.2).

In summary, our key contributions are as follows:
\begin{itemize}
    \item We revisit unified world--action models through safety post-training, showing that augmenting flow-matching imitation with near-expert negative supervision improves challenging-scenario performance without compromising routine driving.
    \item We propose NegWAM, a near-expert negative safety post-training method for WAMs that jointly optimizes the original flow-matching objective and two complementary near-expert negative objectives without introducing additional inference-time overhead.
    \item We introduce two complementary near-expert negative mechanisms. Repulsive anchors push predictions away from locally unsafe trajectories, while scene-conditioned rider tokens enable the unified WAM to determine whether a trajectory is compatible with the current scene.
\end{itemize}

The remainder of this paper is organized as follows. Section~\ref{sec:rela} reviews related work on end-to-end autonomous driving, world--action models, and policy post-training. Section~\ref{sec:method} introduces NegWAM, including its core components and training objective. Section~\ref{sec:exp} evaluates its planning performance and zero-shot closed-loop generalization, with ablation studies and supporting analyses. Section~\ref{sec:conclusion} summarizes the main findings and discusses limitations and future directions.

\section{Literature Review}
\label{sec:rela}

\begin{figure*}[!t]
\centering
\includegraphics[width=\textwidth]
{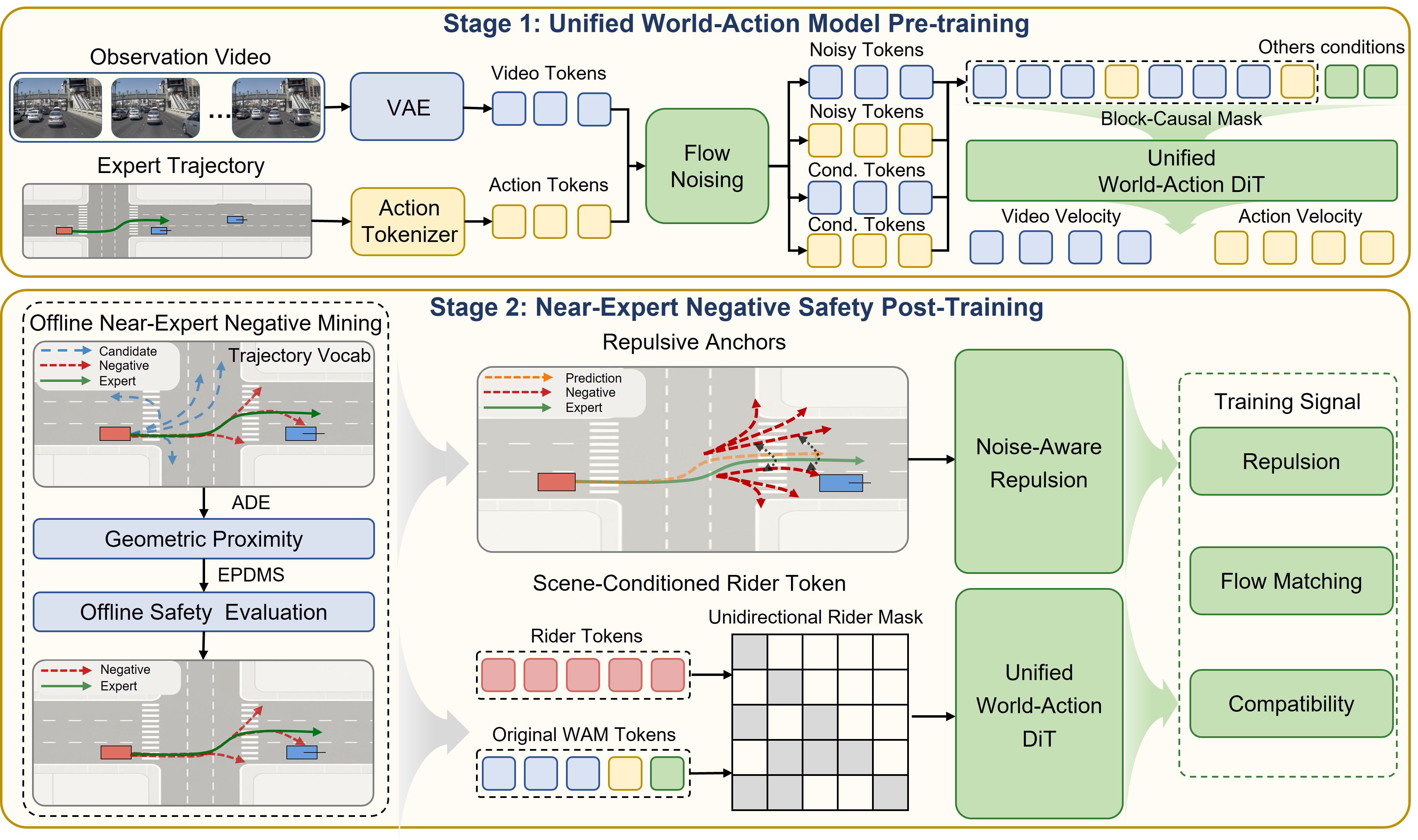}
\caption{
Architecture of our NegWAM.
In Stage 1, we adapt a pretrained world-action DiT into a unified world--action model through flow matching.
In Stage 2, scene-specific hard-negative trajectories are used as both repulsive anchors and trajectory riders for compatibility learning.
The auxiliary rider branch is removed at inference, introducing no additional deployment cost.}
\label{fig:fig2_methods}
\end{figure*}

\subsection{End-to-End Autonomous Driving}
End-to-end autonomous driving maps sensory observations to planned trajectories through a learned policy~\cite{chen2024end-to-end, zhang2026vira}. Joint optimization connects scene understanding with planning and reduces information loss across separately designed modules~\cite{hu2023uniad}. Scaling driving data further improves the coverage of learned behavior~\cite{naumann2025data}, while generative future prediction introduces temporal structure and motion priors into policy learning~\cite{wang2024drivedreamer, zhang2025epona}. These developments strengthen the policy's ability to represent traffic scenes and reproduce expert decisions across diverse driving conditions.

Expert demonstrations nevertheless provide an uneven account of driving behavior: successful maneuvers are abundant, whereas safety-critical failures and recovery interactions are comparatively scarce~\cite{karkus2025beyond, li2026world, yang2026posttraining}. Expert-trajectory supervision identifies a desirable action for the observed scene, but supplies few explicit comparisons with nearby unsafe alternatives. Such alternatives are informative because geometric similarity to an expert trajectory does not establish that an action is safe in the same traffic context~\cite{wang2026beyond, shang2026drivedpo}. Learning to distinguish these alternatives introduces a specific supervision requirement beyond reproducing the demonstrated trajectory.

\subsection{World-Action Models for Driving}
Pretrained video generative models provide temporal and motion priors that can be transferred to action learning~\cite{li2026causal, xia2026drivelaw, shi2026drivewam}. Their role in driving has expanded from supporting scene synthesis and simulation~\cite{yang2026resim, yang2026geniedrive} to providing future representations for trajectory prediction~\cite{zhou2024dino, wang2026drive, wang2026latent}. Recent approaches further couple future visual modeling with action generation, bringing anticipated scene evolution into the policy's prediction process~\cite{xia2026drivelaw, zhou2026drivedreamer, xu2026eponav2}. This progression makes world modeling increasingly relevant to how driving actions are represented and generated.

Recent work~\cite{shi2026drivewam,liu2026driveva} realizes this coupling through shared video--action transformers trained using flow matching. Such architectures allow action supervision to shape the shared world--action representations. Their training objectives, however, focus on fitting demonstrated video--action sequences without explicitly contrasting expert trajectories with undesirable alternatives. This raises the question of how negative trajectories can provide supervision for both action generation and scene--trajectory compatibility during safety post-training.

\subsection{Post-Training for Driving}
Driving policy post-training introduces feedback beyond expert matching through simulated rollouts, reward-based optimization, and trajectory preferences~\cite{karkus2025beyond, yang2026posttraining}. Closed-loop learning uses synthetic simulators such as CARLA~\cite{dosovitskiy2017carla, zhang2021end}, reconstructed scenes, or learned environments to expose policies to the consequences of their actions~\cite{gao2026rad, li2026world}. These mechanisms often overlap: DreamerAD~\cite{yang2026dreamerad} combines latent-world rollouts with learned reward feedback, while WorldRFT~\cite{yang2026worldrft} uses collision-aware rewards for reinforcement fine-tuning. Complementary to policy post-training, TOAD~\cite{xu2026toad} uses a learned trajectory scorer to guide test-time search without retraining the policy.

Preference-based methods express the additional supervision through comparisons between trajectories. TakeAD~\cite{liu2025takead} derives preferences from expert takeover data. DriveDPO~\cite{shang2026drivedpo} combines imitation similarity with rule-based safety scores for policy learning, then applies Safety DPO using human-like but unsafe rejected trajectories. Its preference construction thus retains explicit safety evaluation without requiring a separately trained reward model. BeyondDrive~\cite{wang2026beyond} develops this failure-based perspective through a flow-matching generator, expert-proximity and safety filtering, and a repulsive distance loss defined on predicted trajectories. These studies establish the value of near-expert failures as supervision, with distinct choices of policy objective and negative construction.

For a unified WAM, an unsafe alternative is both a trajectory to avoid and an example of incompatibility between the scene and an action. NegWAM learns these two aspects from the same offline-mined failures: repulsive anchors supervise the generated action trajectory, while rider tokens expose the expert and negative trajectories to the observed scene through shared attention for compatibility learning. The original flow-matching objective remains active throughout post-training. Safety evaluation supplies the offline negative labels; subsequent optimization internalizes this supervision in the WAM, and the auxiliary rider branch is removed at inference.

\section{Methods}
\label{sec:method}

NegWAM post-trains a unified world--action model using scene-specific failures close to expert trajectories. Each retained failure supplies two related learning signals: a motion to avoid during action generation and a scene--trajectory pair to distinguish from the expert. Figure~\ref{fig:fig2_methods} illustrates how repulsive anchors and scene-conditioned rider tokens introduce these signals into the same backbone while retaining its flow-matching objective. 
\subsection{Preliminaries}
\label{sec:preliminaries}

\subsubsection{Unified World--Action Model}
To couple future scene modeling with action generation in one backbone, we adopt a unified world--action architecture~\cite{li2026causal, shi2026drivewam} built on a pretrained video backbone~\cite{wan2025wan} as the base model. A training sample comprises a current camera image $\mathcal{I}^{\mathrm{obs}}$, future visual targets $\mathcal{I}^{\mathrm{fut}}$, an expert trajectory $\tau^+$, and navigation and ego-state conditions including four historical ego poses. A frozen causal video VAE encodes the observed and future frames into the clean visual stream $\mathbf{x}_0^v$. The action tokenizer forms $\mathbf{x}_0^a=E_a(\tau^+)$ by normalizing the expert's eight future $\Delta$-pose steps and prepending a zero conditioning chunk. Prediction masks exclude the observed visual prefix and the conditioning action chunk from the losses. We use $m\in\{v,a\}$ for the two modalities.

For each future chunk and modality, we independently sample Gaussian noise $\boldsymbol{\epsilon}^m$ and a scheduler index, whose shifted noise scale defines $t^m\in[0,1]$. We suppress chunk indices below. On prediction elements, flow matching constructs
\begin{equation}
    \mathbf{x}_t^m
    =
    (1-t^m)\mathbf{x}_0^m
    +
    t^m\boldsymbol{\epsilon}^m,
    \qquad
    \mathbf{v}^m
    :=
    \frac{\partial \mathbf{x}_t^m}{\partial t^m}
    =
    \boldsymbol{\epsilon}^m-\mathbf{x}_0^m.
\end{equation}

Let $\mathbf{t}=(t^v,t^a)$, the shared DiT receives a four-segment sequence
\begin{equation}
    \mathcal{S}_{\mathbf{t}} =
    [\,\mathbf{x}_t^v
    \,\Vert\,\mathbf{x}_c^v
    \,\Vert\,\mathbf{x}_t^a
    \,\Vert\,\mathbf{x}_c^a\,],
\end{equation}
where $\mathbf{x}_c^m$ is the corresponding teacher-forcing conditioning stream. During training, conditioning chunks are exposed only according to their causal temporal positions; future targets are not available as observations at inference. The video conditioning stream may receive noise augmentation, while its observed prefix remains clean. The action conditioning stream uses the zero initial chunk and preceding expert-action chunks. All streams share DiT blocks, with separate modality input and output projections. Under the original block-causal mask $\mathcal{A}_{\mathrm{base}}$, noisy tokens can use preceding conditioning chunks and other noisy tokens in the same temporal block; conditioning tokens retain causal visibility. A finite temporal window limits their history. NegWAM preserves this mask for all original tokens.

\subsubsection{Flow Matching}
Flow matching~\cite{liu2022flow,lipman2022flow,tong2023improving} learns a velocity field along a continuous interpolation between the data distribution and Gaussian noise. The shared DiT jointly predicts the video and action velocity fields:
\begin{equation}
    (\hat{\mathbf{v}}_\theta^v,\hat{\mathbf{v}}_\theta^a)
    =
    F_\theta(\mathcal{S}_{\mathbf{t}};\mathbf{c}),
\end{equation}
where $\mathbf{c}$ contains command guidance and ego-state
conditions. The original training objective is
\begin{equation}
    \mathcal{L}_{\mathrm{FM}}
    =
    \sum_{m\in\{v,a\}}
    \mathbb{E}\!\left[
    w_m(t^m)
    \frac{
    \left\|
    \mathbf{M}^m\odot
    (\hat{\mathbf{v}}_\theta^m-\mathbf{v}^m)
    \right\|_2^2
    }{
    \|\mathbf{M}^m\|_1
    }
    \right],
\end{equation}
where $\mathbf{M}^m$ selects valid future prediction elements and $w_m(t^m)$ is the scheduler's normalized flow-matching weight. The scheduler grid, sampling rule, and weight function are specified in the supplementary implementation details. Under this parameterization, the clean action estimate and its decoded trajectory are
\begin{equation}
    \label{eq:clean_action_estimate}
    \hat{\mathbf{x}}_0^a
    =
    \mathbf{x}_t^a-t^a\hat{\mathbf{v}}_\theta^a,
    \qquad
    \hat{\tau}
    =
    D_a(\hat{\mathbf{x}}_0^a),
\end{equation}
where $D_a$ selects the future action chunk, denormalizes its $\Delta$-pose channels, and cumulatively integrates them into ego-frame poses. Our method preserves this flow-matching pathway and
appends trajectory riders as additional action-token groups processed by the shared DiT.

\subsection{Near-Expert Negative Safety Post-Training}

\subsubsection{Near-Expert Negative Trajectory Construction}
Directly sampling failure trajectories from the current policy can be
ineffective under strong scene conditioning, consistent with prior
observations of limited sampling diversity~\cite{yang2026posttraining}.
In a diagnostic probe using the same sampling configuration as the main experiments, the average pairwise ADE among eight trajectories generated for the same scene is only $0.02\,\mathrm{m}$, indicating limited diversity in this probe. We therefore mine negative trajectories from the 16K dense trajectory vocabulary of
GTRS~\cite{li2025gtrs}. Each vocabulary trajectory $\bar\tau_j$ contains 40 ego-frame poses over a $4\,\mathrm{s}$ horizon. The subsampling operator $S$ takes every fifth pose, corresponding to $0.5\,\mathrm{s}$ intervals, to obtain the eight-step trajectory $\tau_{i,j}=S(\bar\tau_j)$ aligned with the expert $\tau_i^+$. We measure their distance using
the Average Displacement Error (ADE) over their planar positions $\mathbf{p}^t=(x^t,y^t)$:
\begin{equation}
    a_{i,j}
    =
    \operatorname{ADE}(\tau_{i,j},\tau_i^+)
    =
    \frac{1}{T}
    \sum_{t=1}^{T}
    \left\|
        \mathbf{p}_{i,j}^{t}-\mathbf{p}_i^{+,t}
    \right\|_2 .
\end{equation}
Algorithm~\ref{algo:pipeline} separates offline failure mining from the preparation of training negatives. For each training scene, we first retain vocabulary candidates with $a_{i,j}<1.0\,\mathrm{m}$. We evaluate their original 40-step trajectories using the scene's NAVSIM v2 metric cache and log-replay traffic. All candidates with zero EPDMS~\cite{cao2025navsimv2} are cached as evaluator-defined failures, forming a scene-specific pool $\mathcal{H}_i$.

At each training iteration, the cached pool $\mathcal{H}_i$ of the corresponding scene is subsampled to eight steps per trajectory, and the expert-relative ADEs are recomputed on the subsampled poses. We retain candidates with ADE at least $0.3\,\mathrm{m}$, giving the final range $[0.3,1.0)\,\mathrm{m}$. From these candidates, we randomly select at most $K=16$ without replacement and zero-pad the trajectory and ADE tensors. A validity mask $M_{i,k}^{-}$ distinguishes selected negatives from padding. The same selected subset supplies both hard-negative objectives for that update; different reads may select different subsets from the cached pool. Samples without valid negatives contribute only the flow-matching loss. EPDMS is queried only during offline mining.

\begin{algorithm}[!t]
\caption{Near-Expert Negative Trajectory Construction}
\label{algo:pipeline}
\small
\begin{algorithmic}[1]
\REQUIRE Training scenes $s_i$ with expert trajectories $\tau_i^+$; 40-step vocabulary $\mathcal{V}=\{\bar\tau_j\}$; offline EPDMS evaluator $\mathcal{E}$
\REQUIRE $d_{\min}=0.3\,\mathrm{m}$, $d_{\max}=1.0\,\mathrm{m}$, $K=16$; eight-step subsampling operator $S$
\ENSURE Cached failure pools $\mathcal{H}_i$ and per-read negatives, expert ADEs, and validity masks
\STATE \textbf{Offline scene-conditioned failure mining}
\FOR{each training scene $s_i$}
    \STATE Compute expert ADEs $a_{i,j}$ on $S(\bar\tau_j)$ for all candidates
    \STATE $\mathcal{C}_i\gets\{\bar\tau_j\in\mathcal{V}:a_{i,j}<d_{\max}\}$
    \STATE $\mathcal{H}_i\gets\{\bar\tau\in\mathcal{C}_i:\mathcal{E}(s_i,\bar\tau)=0\}$
    \STATE Cache all trajectories in $\mathcal{H}_i$ at their original 40 steps
\ENDFOR
\STATE \textbf{Preparing a training negative set}
\FOR{each training-sample read for scene $i$}
    \STATE Load $\mathcal{H}_i$; subsample each trajectory to eight steps using $S$
    \STATE Compute expert ADE $a_{i,j}$ for each loaded trajectory
    \STATE $\mathcal{J}_i\gets\{j:a_{i,j}\ge d_{\min}\}$; $q_i\gets\min(K,|\mathcal{J}_i|)$
    \STATE Draw $q_i$ indices from $\mathcal{J}_i$ uniformly without replacement
    \STATE Initialize $K$ trajectory, ADE, and validity entries to zero
    \STATE Fill the first $q_i$ entries with selected trajectories and ADEs
    \STATE Set $M_{i,k}^-=1$ for $k\le q_i$; all padding remains masked
    \STATE Output $\{(\tau_{i,k}^-,a_{i,k},M_{i,k}^-)\}_{k=1}^{K}$
\ENDFOR
\end{algorithmic}
\end{algorithm}

This procedure yields at least one valid hard negative for $85.8\,\%$ of the training samples. Across all valid, unpadded negatives, the median ADE to the expert trajectory is $0.80\,\mathrm{m}$, and approximately $9\,\%$ fall within $[0.3,0.5]\,\mathrm{m}$.

\subsubsection{Noise-Aware Anchor Repulsion}
We treat the selected negative trajectories as fixed repulsive anchors
in trajectory space. Let $d_{i,k}$ denote the ADE between the one-step
decoded trajectory $\hat{\tau}_i$ from
Eq.~\eqref{eq:clean_action_estimate} and the $k$-th negative
$\tau_{i,k}^{-}$, and let $a_{i,k}$ denote the ADE between this negative
and the expert trajectory. A fixed margin cannot accommodate negatives
at different distances: a small margin leaves many negatives inactive,
whereas a large one may conflict with the imitation objective. We
therefore define the repulsion loss using a distance-adaptive margin:
\begin{equation}
    \label{eq:repulsion_hinge}
    \ell_{i,k}^{\mathrm{rep}}
    =
    \left[\alpha a_{i,k}-d_{i,k}\right]_+,
    \qquad
    0<\alpha\leq1.
\end{equation}
We use $\alpha=0.9$. The loss is active within the distance-dependent neighborhood of a negative anchor. At the exact expert trajectory, $d_{i,k}=a_{i,k}$, so this repulsion term is zero for $\alpha\leq1$.

Since one-step trajectory estimates are less reliable at high noise
levels, we weight the repulsion term by $1-t_i^{a}$ and
average it over valid negatives:
\begin{equation}
    \label{eq:repulsion_loss}
    \mathcal{L}_{\mathrm{rep}}
    =
    \mathbb{E}_{i\sim\mathcal{B}}
    \left[
        (1-t_i^a)
        \frac{
            \sum_k M_{i,k}^{-}
            \left[\alpha a_{i,k}-d_{i,k}\right]_+
        }{
            \max(1,\sum_k M_{i,k}^{-})
        }
    \right],
\end{equation}
where $\mathcal{B}$ is the training minibatch. A sample without valid negatives contributes zero. Differentiating the reconstruction in Eq.~\eqref{eq:clean_action_estimate} introduces a factor $t_i^a$; together with the loss weight, this gives a $t_i^a(1-t_i^a)$ prefactor that vanishes at the noise endpoints. The actual gradient also depends on the trajectory decoder and active hinge terms. Repulsion reuses the original action output without another backbone forward pass and defines its supervision through trajectory distances.

\subsubsection{Scene-Conditioned Rider Tokens}
\begin{figure}[!t]
    \centering
    \includegraphics[width=\linewidth]{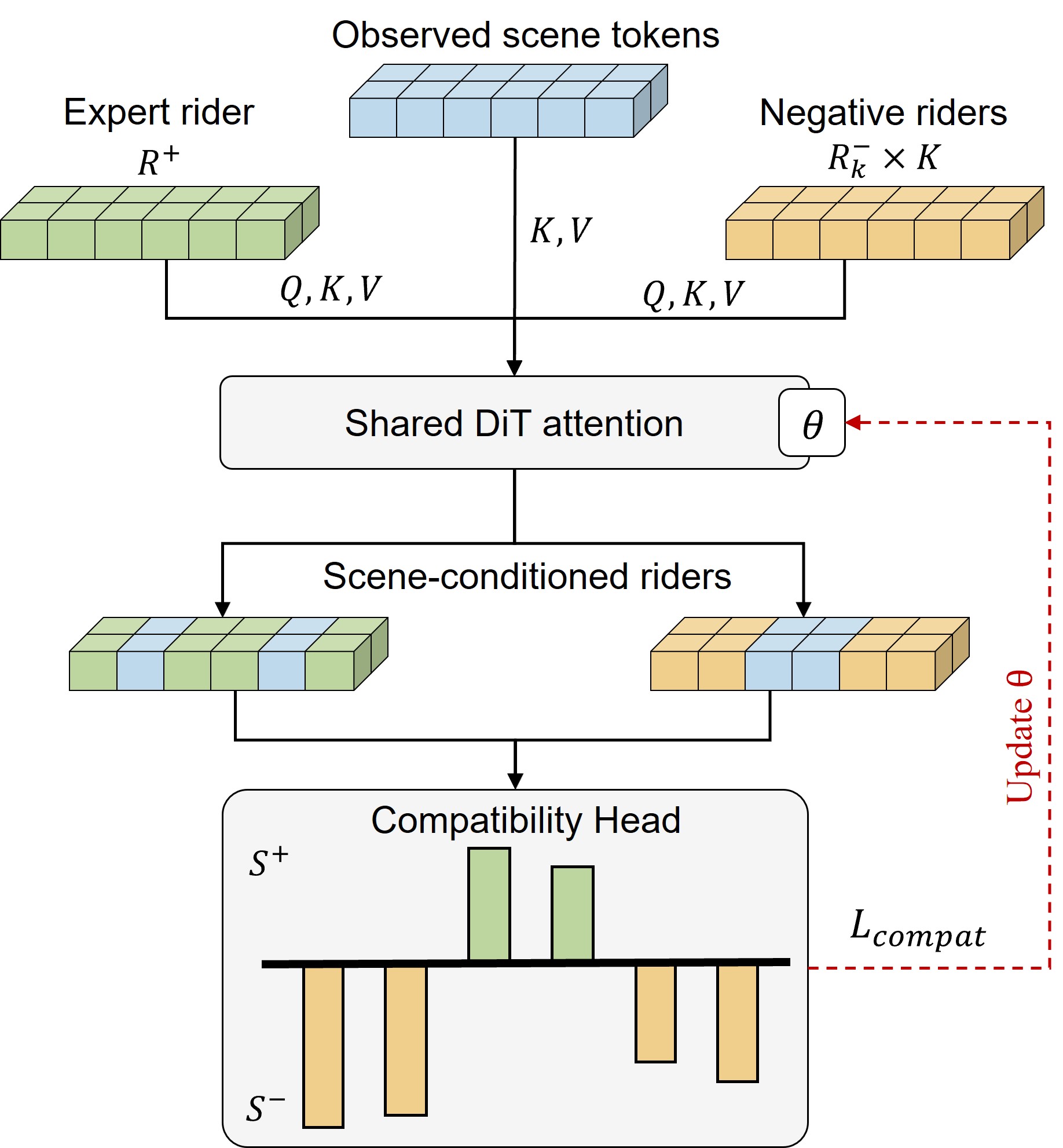}
    \caption{Scene-conditioned rider tokens for scene--trajectory compatibility learning. Expert and negative riders attend to observed scene tokens through shared DiT attention, and their compatibility scores define a ranking loss that updates the shared backbone.}
    \label{fig:rider}
\end{figure}

The same failures also define scene--trajectory comparisons. We encode the expert and each negative as a separate auxiliary rider group $\mathcal{R}_g$, using the shared action embedding and the same rotary positional encoding grid. Each group contains a zero conditioning chunk and an eight-step trajectory chunk, giving 16 action tokens. Riders are clean, with no diffusion noise and a flow-time input $t=0$. The expert occupies group zero; the remaining groups represent the negatives. Figure~\ref{fig:rider} illustrates the shared attention and scoring pathway.

The riders access the scene through shared attention while remaining invisible to the original streams. Let $\mathbf{x}_c^{v,\mathrm{obs}}$ contain the clean visual conditioning tokens of the current observed frame. At every DiT block, the attention visibility is
\begin{equation}
    \label{eq:rider_mask}
    \mathcal{A}(p,q)
    =
    \begin{cases}
        \mathcal{A}_{\mathrm{base}}(p,q),
        & p,q\in\mathcal{S}_{\mathbf{t}},\\
        1,
        & p\in\mathcal{R}_g,\;
          q\in\mathbf{x}_c^{v,\mathrm{obs}}\cup\mathcal{R}_g,\\
        0,
        & \text{otherwise}.
    \end{cases}
\end{equation}
Thus, each rider sees only the current observed visual tokens and its own trajectory group. Future target tokens are excluded from these keys, and the observed tokens retain their original causal visibility. The original sequence cannot attend to riders, so appending them preserves its forward attention path at fixed backbone weights. Matching the expert and negative groups' tokenization, positions, and visibility reduces structural differences in their scoring inputs.

We average the output features over all 16 tokens in each group and apply a shared compatibility head $h_\phi$: a linear projection to 256 dimensions, SiLU, and a scalar linear output. Zero-initializing the final layer's weights and bias gives identical initial scores and zero initial compatibility gradient to the backbone; this gradient path becomes active as the head's output weights are updated. Let $s_i^+$ and $s_{i,k}^-$ be the expert and negative scores. For $\mathcal{B}_+=\{i\in\mathcal{B}:\sum_k M_{i,k}^->0\}$, we optimize
\begin{equation}
    \label{eq:compatibility_loss}
    \begin{aligned}
        \mathcal{L}_{\mathrm{compat}}
        &=-\mathbb{E}_{\mathcal{B}_+}
        \log\frac{\exp(s_i^+/\tau_s)}{Z_i},\\
        Z_i&=\exp(s_i^+/\tau_s)
        +\sum_k M_{i,k}^{-}\exp(s_{i,k}^-/\tau_s).
    \end{aligned}
\end{equation}
where $\tau_s=0.1$. The expectation denotes the mean over samples with valid negatives; we set $\mathcal{L}_{\mathrm{compat}}=0$ when $\mathcal{B}_+$ is empty. This objective ranks the expert above valid scene-specific negatives. Its gradients propagate through the shared attention projections, adding an explicit scene--trajectory ranking objective to the backbone. Rider groups and the compatibility head are used only during post-training.

\subsection{Training Objective}
\label{sec:training_objective}

Our training procedure consists of two consecutive stages. In the
first stage, we initialize the model from the pretrained video
generative model Wan2.2-TI2V-5B~\cite{wan2025wan} and adapt it to autonomous driving by following the unified world--action flow-matching paradigm of
DriveWAM~\cite{shi2026drivewam}. This stage optimizes only the flow-matching objective defined in Section~\ref{sec:preliminaries}, enabling the model to jointly learn future visual dynamics and expert trajectory prediction. The resulting model serves as the base WAM checkpoint shared by all subsequent
experiments.

In the second stage, we initialize from this shared checkpoint and
perform hard-negative safety post-training. We retain the original
flow-matching objective as supervision for visual dynamics and expert actions, while introducing repulsive trajectory
supervision and scene-conditioned compatibility learning. The overall
post-training objective is
\begin{equation}
    \label{eq:post_training_loss}
    \mathcal{L}_{\mathrm{post}}
    =
    \mathcal{L}_{\mathrm{FM}}
    +
    \lambda_{\mathrm{rep}}\mathcal{L}_{\mathrm{rep}}
    +
    \lambda_{\mathrm{compat}}\mathcal{L}_{\mathrm{compat}},
\end{equation}
where $\mathcal{L}_{\mathrm{rep}}$ pushes predictions away from unsafe
neighbors in trajectory space, and $\mathcal{L}_{\mathrm{compat}}$
learns scene--trajectory compatibility through rider tokens.
$\lambda_{\mathrm{rep}}$ and $\lambda_{\mathrm{compat}}$ control the
contributions of the two hard-negative objectives. All post-training variants start from the same first-stage checkpoint, controlling for differences in base-model initialization.

At inference, we discard the rider groups and compatibility head and use the original world--action generation pathway with the updated backbone. The model encodes the available observations, generates future visual latents with three Euler updates, and then generates the eight-step action chunk with ten Euler updates. These are integration-step counts rather than total backbone forward-call counts. The action decoder converts the generated tokens to the final trajectory; no hard-negative candidates or evaluator are used at this stage.

\section{Experiments and Results}
\label{sec:exp}
\begin{table*}[!t]
    \caption{Performance comparison on the NAVSIM v2 \textnormal{\textit{navhard}} benchmark.
    S1 denotes evaluation on the initial real-world scenes, while S2 denotes evaluation on the pre-generated reactive follow-up scenes.
    All results are reported using the corrected official NAVSIM v2.2 evaluation protocol. MV and SV denote multi-view and single-view camera inputs, respectively. DriveFuture\textsuperscript{$\dagger$} denotes its variant without the GTRS-Dense scorer; the full system reports 55.5 with the GTRS-Dense scorer~\cite{li2025gtrs}.}
    \label{tab:table2_navhard}
    \centering
    \resizebox{\linewidth}{!}{
    \begin{tabular}{l|l|l|l|llllllllll}
    \toprule
    \textbf{Methods} 
    & {\textbf{Sensor}}
    & {\textbf{Stage}} 
    & {\textbf{NC} $\uparrow$} 
    & {\textbf{DAC} $\uparrow$}  
    & {\textbf{DDC} $\uparrow$}     
    & {\textbf{TLC} $\uparrow$}     
    & {\textbf{EP}  $\uparrow$}
    & {\textbf{TTC} $\uparrow$}  
    & {\textbf{LK} $\uparrow$}   
    & {\textbf{HC} $\uparrow$}    
    & {\textbf{EC} $\uparrow$}       
    & {\textbf{EPDMS} $\uparrow$}                      
    \\ 
    \midrule
    
    & {}
    & {S1}   
    & {94.4} 
    & {98.8} 
    & {100}  
    & {99.5}
    & {100} 
    & {93.5}   
    & {99.3} 
    & {87.7} 
    & {36.0}     
    & {}    
    
    \\
    \multirow{-2.5}{*}{PDM-Closed} 
    & \multicolumn{1}{c|}{\multirow{-2.5}{*}{-}}
    & {S2}
    & {90.5} 
    & {90.6}
    & {95.4} 
    & {98.4}
    & {100}
    & {86.6}
    & {74.2} 
    & {91.9}
    & {29.7}  
    & {\multirow{-2}{*}{56.6}}    \\

    \midrule
    \hline
    \multicolumn{13}{l}{\cellcolor[HTML]{DCEAF7}\textit{\textbf{E2E-based Methods}}}
    \\
    & {}
    & {S1}
    & {96.0}
    & {79.7}
    & {97.4}
    & {99.5}
    & {81.3}
    & {93.1}
    & {90.8}
    & {96.8}
    & {73.8}
    & {} \\

    \multirow{-2.5}{*}{DiffusionDrive~\cite{liao2025diffusiondrive}}
    & \multicolumn{1}{c|}{\multirow{-2.5}{*}{MV}}
    & {S2}
    & {82.1}
    & {72.2}
    & {88.5}
    & {98.7}
    & {85.1}
    & {78.8}
    & {49.2}
    & {89.3}
    & {71.2}
    & {\multirow{-2}{*}{24.2}} \\

    & {}
    & {S1}
    & {96.6}
    & {80.5}
    & {96.3}
    & {99.3}
    & {82.3}
    & {94.9}
    & {91.5}
    & {97.7}
    & {67.8}
    & {} \\

    \multirow{-2.5}{*}{GuideFlow~\cite{liu2026guideflow}}
    & \multicolumn{1}{c|}{\multirow{-2.5}{*}{MV}}
    & {S2}
    & {87.3}
    & {76.7}
    & {88.8}
    & {99.2}
    & {84.3}
    & {85.1}
    & {49.7}
    & {93.1}
    & {44.2}
    & {\multirow{-2}{*}{27.1}} \\
    
    \midrule
    \hline
    \multicolumn{13}{l}{\cellcolor[HTML]{DCEAF7}\textit{\textbf{VLA-based Methods}}}
    \\
    & {}
    & {S1}
    & {96.8}
    & {83.3}
    & {99.0}
    & {99.6}
    & {84.6}
    & {95.3}
    & {96.4}
    & {97.6}
    & {78.2}
    & {}  \\

    \multirow{-2.5}{*}{DriveVLA-W0~\cite{li2025drivevla}} 
    & \multicolumn{1}{c|}{\multirow{-2.5}{*}{SV}}
    & {S2}
    & {76.8}
    & {64.3}
    & {79.9}
    & {98.3}
    & {89.2}
    & {75.0}
    & {46.8}
    & {95.8}
    & {53.1}
    & {\multirow{-2}{*}{24.4}} \\

    & {}
    & {S1}
    & {98.4}
    & {94.3}
    & {97.8}
    & {99.9}
    & {85.7}
    & {97.2}
    & {94.2}
    & {97.6}
    & {72.1}
    & {} \\

    \multirow{-2.5}{*}{SpanVLA~\cite{zhou2026spanvla}}
    & \multicolumn{1}{c|}{\multirow{-2.5}{*}{MV}}
    & {S2}
    & {86.9}
    & {84.3}
    & {87.1}
    & {98.2}
    & {85.5}
    & {82.7}
    & {62.3}
    & {96.8}
    & {67.4}
    & {\multirow{-2}{*}{40.1}} \\

    \midrule
    \hline
    \multicolumn{13}{l}{\cellcolor[HTML]{DCEAF7}\textit{\textbf{World-Model-based Methods}}}
    \\

    & {}
    & {S1}   
    & {97.1} 
    & {88.7} 
    & {99.2}  
    & {99.6}
    & {84.2} 
    & {97.1}   
    & {96.2} 
    & {97.8} 
    & {69.8}    
    & {}    
    
    \\
    \multirow{-2.5}{*}{DriveLAW~\cite{xia2026drivelaw}} 
    & \multicolumn{1}{c|}{\multirow{-2.5}{*}{SV}}
    & {S2}
    & {82.3} 
    & {67.9}
    & {83.7} 
    & {98.1}
    & {84.7}
    & {78.5}
    & {46.7} 
    & {96.3}
    & {60.6}  
    & {\multirow{-2}{*}{30.4}}  \\

    & {}
    & {S1}
    & {96.3}
    & {87.8}
    & {98.2}
    & {99.6}
    & {83.1}
    & {96.9}
    & {94.9}
    & {97.6}
    & {76.9}
    & {} \\

    \multirow{-2.5}{*}{DriveFuture\textsuperscript{$\dagger$}~\cite{hong2026drivefuture}}
    & \multicolumn{1}{c|}{\multirow{-2.5}{*}{MV}}
    & {S2}
    & {82.3}
    & {78.8}
    & {88.1}
    & {98.4}
    & {83.6}
    & {79.6}
    & {47.6}
    & {97.0}
    & {75.9}
    & {\multirow{-2}{*}{34.6}} \\

    & {} 
    & {S1}
    & {97.3}
    & {90.7}
    & {99.4}
    & {100}
    & {83.3}
    & {97.3}
    & {97.3}
    & {97.6}
    & {60.9}
    & {} \\

    \multirow{-2.5}{*}{EponaV2~\cite{xu2026eponav2}}
    & \multicolumn{1}{c|}{\multirow{-2.5}{*}{SV}}
    & {S2}
    & {83.6}
    & {78.0}
    & {88.0}
    & {98.9}
    & {86.0}
    & {80.3}
    & {50.1}
    & {96.1}
    & {52.0}
    & {\multirow{-2}{*}{36.1}}  \\
    
    & {}
    & {S1}   
    & {96.8} 
    & {88.7} 
    & {99.1}  
    & {99.6}
    & {95.9} 
    & {95.6}   
    & {95.6} 
    & {97.8} 
    & {80.4}     
    & {}

    \\
    \multirow{-2.5}{*}{DriveWAM~\cite{shi2026drivewam}} 
    & \multicolumn{1}{c|}{\multirow{-2.5}{*}{SV}}
    & {S2}
    & {79.5} 
    & {75.8}
    & {87.8} 
    & {98.1}
    & {96.5}
    & {75.7}
    & {48.7} 
    & {95.8}
    & {73.7}  
    & {\multirow{-2}{*}{37.7}}  \\

    & {}
    & {S1}   
    & {96.6} 
    & {93.1} 
    & {98.8}  
    & {99.6}
    & {96.9} 
    & {95.8}   
    & {96.9} 
    & {97.8} 
    & {78.7}     
    & {} 

    \\
    \multirow{-2.5}{*}{BeyondDrive~\cite{wang2026beyond}} 
    & \multicolumn{1}{c|}{\multirow{-2.5}{*}{SV}}
    & {S2}
    & {81.2} 
    & {77.7}
    & {85.6} 
    & {98.2}
    & {97.0}
    & {77.7}
    & {47.8} 
    & {95.4}
    & {72.2}  
    & {\multirow{-2}{*}{38.8}}  \\

    \rowcolor[HTML]{FEF0C7}
    & {}
    & {S1}
    & {96.6}
    & {94.4}
    & {98.8}
    & {99.6}
    & {95.8}
    & {96.0}
    & {96.4}
    & {97.8}
    & {79.6}
    & {} \\

    \rowcolor[HTML]{FEF0C7}
    \multirow{-2}{*}{\textbf{Ours}}
    & \multicolumn{1}{c|}{\multirow{-2}{*}{SV}}
    & {S2}
    & {79.2}
    & {78.0}
    & {86.4}
    & {98.1}
    & {96.3}
    & {75.3}
    & {48.4}
    & {95.7}
    & {73.7}
    & {\multirow{-2}{*}{\textbf{40.4}}} \\

    \bottomrule
    \end{tabular}}
\end{table*}

In this section, we investigate the following questions:
\begin{itemize}
    \item Does NegWAM improve planning performance
    on both challenging and routine benchmarks, and do these gains
    transfer to zero-shot closed-loop evaluation?
    (Section~\ref{exp:main})

    \item What are the individual and combined contributions of
    repulsive anchors and scene-conditioned rider tokens,
    and how sensitive is performance to the post-training budget
    and hyperparameters?
    (Section~\ref{exp:abla})

    \item What qualitative driving behaviors are observed in
    trajectory comparisons and closed-loop rollouts?
    (Section~\ref{exp:quali})

    \item What inference cost accompanies
    the planning improvements?
    (Section~\ref{exp:efficiency})
\end{itemize}

\subsection{Experimental Setup}

\subsubsection{Datasets and Benchmarks}
NAVSIM~\cite{dauner2024navsim} is a real-world autonomous
driving dataset built upon OpenScene~\cite{openscene2023} and
nuPlan~\cite{karnchanachari2024towards}, comprising 103K training
samples. We train all models on \textit{navtrain} and evaluate them on
\textit{navtest} and \textit{navhard}. To further assess zero-shot closed-loop
generalization, we directly evaluate the same checkpoints on
HUGSIM~\cite{zhou2025hugsim}.

\subsubsection{Evaluation Metrics}
\label{exp:metrics}
NAVSIM v2. Both \textit{navtest} and \textit{navhard} report the Extended Predictive Driver Model Score (EPDMS)~\cite{li2025hydramdpplusplus}, which combines multiplicative penalties with a weighted average of sub-metric scores:
\begin{equation}
    \mathrm{EPDMS}
    =
    \left(\prod_{m\in\mathcal{M}_{\mathrm{pen}}} S_m\right)
    \frac{\sum_{m\in\mathcal{M}_{\mathrm{avg}}} w_m S_m}
         {\sum_{m\in\mathcal{M}_{\mathrm{avg}}} w_m},
    \label{eq:epdms_metric}
\end{equation}
where $S_m$ and $w_m$ denote the score and weight of sub-metric $m$, respectively. The multiplicative penalty set $\mathcal{M}_{\mathrm{pen}}$ comprises No At-Fault Collisions (NC), Drivable Area Compliance (DAC), Driving Direction Compliance (DDC), and Traffic Light Compliance (TLC). The weighted-average set $\mathcal{M}_{\mathrm{avg}}$ includes Time-to-Collision (TTC), Ego Progress (EP), Lane Keeping (LK), History Comfort (HC), and Extended Comfort (EC). For \textit{navhard}, scoring is further adapted to the two-stage setting through cross-stage aggregation, reactive traffic simulation, and the exemption of penalties when the corresponding expert rollout exhibits the same failure. All results use the corrected official NAVSIM v2.2 evaluation protocol, which fixes an inconsistency in the human-penalty filtering of the aggregated score~\cite{cao2025navsimv2}. All compared methods follow the same corrected evaluation protocol.

HUGSIM. We report per-level and overall Route Completion (RC) and HD-Score. For an episode of length $T$, HD-Score combines RC with No Collision (NC), Drivable Area Compliance (DAC), Time to Collision (TTC), and Comfort (COM):
\begin{equation}
    \begin{aligned}
        \mathrm{HD\text{-}Score}
        &= \frac{\mathrm{RC}}{T}\sum_{t=1}^{T}
        \Biggl[\mathrm{NC}_{t}\,\mathrm{DAC}_{t} \\
        &\qquad\times
        \frac{5\,\mathrm{TTC}_{t}+2\,\mathrm{COM}_{t}}{7}\Biggr],
    \end{aligned}
    \label{eq:hugsim_hd_score}
\end{equation}
where $t$ indexes the simulation time steps. Higher EPDMS, RC, and HD-Score indicate better planning performance.

\subsubsection{Implementation Details}
Our model uses one current camera image at a resolution of $256\times448$, together with four historical ego poses, while the action branch predicts a $4\,\mathrm{s}$ future trajectory. We train the model on \textit{navtrain} in two stages. Following the unified world--action flow-matching paradigm of DriveWAM~\cite{shi2026drivewam}, we initialize the video backbone from Wan2.2-TI2V-5B~\cite{wan2025wan} and perform 100K steps of WAM pretraining to jointly learn future visual dynamics and expert trajectory prediction. Starting from this checkpoint, we conduct 10K steps of hard-negative safety post-training. We retain the original flow-matching objective to preserve the learned world dynamics and expert-driving prior, while introducing the proposed repulsive-anchor and scene-conditioned rider-token objectives. The auxiliary rider branch is removed at inference, and measured inference efficiency is reported in Section~\ref{exp:efficiency}. Complete model configurations, optimization settings, and training costs are provided in the supplementary material.

For the BeyondDrive~\cite{wang2026beyond} baseline in Tables~\ref{tab:table2_navhard} and~\ref{tab:table1_navtest}, we port the complete method, rather than only its repulsive loss, to our WAM architecture, strictly following the authors' original source code and using BeyondDrive's original negative samples. The reported results are obtained from our own experiments with this implementation and are not taken from the original paper.

\subsection{Planning Performance and Generalization}
\label{exp:main}

\subsubsection{NAVSIM v2}
On \textit{navhard}, NegWAM achieves 40.4 EPDMS, surpassing the BeyondDrive~\cite{wang2026beyond} baseline by 1.6 points, corresponding to a 4.1\,\% relative improvement, and establishing state-of-the-art performance among WAM without an external trajectory scorer (Table~\ref{tab:table2_navhard}). On \textit{navtest}, NegWAM further achieves 93.3 EPDMS, outperforming DriveWAM~\cite{shi2026drivewam} by 1.1 points and establishing a new state of the art among the compared learning-based methods (Table~\ref{tab:table1_navtest}). The consistent improvements across both benchmarks show that the proposed negative-aware post-training strategy enhances robustness in challenging scenarios while preserving, and even improving, the general driving capability acquired by the pretrained WAM.

\begin{table}[!t]
    \caption{Performance comparison on the NAVSIM v2 \textnormal{\textit{navtest}} benchmark.
    All results are reported using the corrected official NAVSIM v2.2 evaluation protocol.
    MV and SV denote multi-view and single-view camera inputs, respectively.}
    \label{tab:table1_navtest}
    \centering
	\resizebox{\linewidth}{!}{
		\begin{tabular}{l|c|ccc}
			\toprule
			\textbf{Methods}  
			&{\textbf{Sensor}}
            &{\textbf{NC} $\uparrow$}
            &{\textbf{DAC} $\uparrow$}
			& \textbf{EPDMS} $\uparrow$ \\ 

            \midrule
            Human agent
            & {-}
            & {100.0}
            & {100.0}
            & {97.9} \\
            \hline
			\multicolumn{5}{l}{\cellcolor[HTML]{DCEAF7} \textit{\textbf{E2E-based Methods}}} 
			\\
            DiffusionDrive~\cite{liao2025diffusiondrive}
            & {MV}
            & {98.2}
            & {95.9}
            & {84.5} \\

            DiffusionDriveV2~\cite{zou2025diffusiondrivev2}
            & {MV}
            & {97.7}
            & {96.6}
            & {87.5} \\

            \hline
			\multicolumn{5}{l}{\cellcolor[HTML]{DCEAF7} \textit{\textbf{VLA-based Methods}}} 
			\\
            Recogdrive~\cite{li2025recogdrive}
            & {MV}
            & {98.3}
            & {95.2}
            & {83.6} \\

            DriveVLA-W0~\cite{li2025drivevla}
            & {SV}
            & {98.5}
            & {99.1}
            & {86.1} \\

            DriveWorld-VLA~\cite{liu2026driveworld}
            & {MV}
            & {98.6}
            & {99.1}
            & {86.8} \\

            \hline
			\multicolumn{5}{l}{\cellcolor[HTML]{DCEAF7} \textit{\textbf{World-Model-based Methods}}} 
			\\
            Latent-WAM~\cite{wang2026latent}
            & \multicolumn{1}{c|}{MV}
            & {98.1}
            & {97.3}
            & {89.3} \\

            DriveFuture~\cite{hong2026drivefuture}
            & \multicolumn{1}{c|}{MV}
            & {98.8}
            & {99.1}
            & {89.9} \\
            
			DriveLAW~\cite{xia2026drivelaw}
			& \multicolumn{1}{c|}{SV} 
            & {98.8}
            & {96.9}
			& {88.9}  \\ 

            Discrete-WAM~\cite{yao2026discrete}
            & \multicolumn{1}{c|}{SV}
            & {98.5}
            & {98.2}
            & {90.4} \\
			
            DriveWAM~\cite{shi2026drivewam}
            & \multicolumn{1}{c|}{SV}
            & {98.4}
            & {96.7}
            & {92.2} \\

            BeyondDrive~\cite{wang2026beyond}
            & {SV}
            & {98.2}
            & {97.7}
            & {92.9}
            \\

            \rowcolor[HTML]{FEF0C7}
            \textbf{Ours}
            & \multicolumn{1}{c|}{SV}
            & {98.1}
            & {98.4}
            & {\textbf{93.3}} \\
            
			\bottomrule
	\end{tabular}}
\end{table}

\subsubsection{Zero-shot Generalization to HUGSIM}
Under zero-shot evaluation on HUGSIM~\cite{zhou2025hugsim}, NegWAM improves RC and HD-Score by 3.4\,\% and 12.5\,\% relative to DriveWAM~\cite{shi2026drivewam}, reaching 45.1 and 30.6, respectively (Table~\ref{tab:hugsim_zero_shot}). The gains are concentrated in the Easy settings, while performance on Hard remains nearly unchanged and both metrics decrease on Extreme. These results show improved aggregate closed-loop performance without HUGSIM-specific fine-tuning, with transfer varying across scenario difficulty.

\begin{table}[!htbp]
\centering\small
\caption{Zero-shot performance on the HUGSIM benchmark.
E/M/H/X denote easy, medium, hard, and extreme scenarios.
Higher is better for both route completion (RC) and HD-Score.}
\label{tab:hugsim_zero_shot}
\setlength{\tabcolsep}{3pt}
\resizebox{\linewidth}{!}{%
\begin{tabular}{@{}lrrrrrrrrrr@{}}
\toprule
& \multicolumn{5}{c}{RC $\uparrow$}
& \multicolumn{5}{c}{HD-Score $\uparrow$} \\
\cmidrule(lr){2-6}\cmidrule(lr){7-11}
Method & E & M & H & X & Overall
       & E & M & H & X & Overall \\
\midrule
VAD~\cite{jiang2023vad}
& 38.7 & 27.0 & 25.5 & 23.0 & 27.9
& 24.3 & 9.9 & 10.4 & 8.2 & 12.3 \\
LTF~\cite{dauner2024navsim}
& 68.4 & 40.7 & 36.9 & 25.5 & 41.4
& 52.8 & 24.6 & 19.8 & 8.1 & 24.8 \\
DriveWAM~\cite{shi2026drivewam}
& 76.6 & 33.8 & 33.9 & 30.2 & 43.6
& 64.0 & 16.1 & 16.1 & 12.6 & 27.2 \\
UniAD~\cite{hu2023uniad}
& 58.6 & 41.2 & 40.4 & 26.0 & 40.6
& 48.7 & 29.5 & 27.3 & 14.3 & 28.9 \\
\midrule
\rowcolor[HTML]{FEF0C7}
\textbf{NegWAM (Ours)}
& 90.3 & 32.8 & 31.2 & 26.1 & \textbf{45.1}
& 81.7 & 15.8 & 14.2 & 10.8 & \textbf{30.6} \\
\bottomrule
\end{tabular}%
}
\end{table}

\subsection{Analysis of Near-Expert Negative Supervision}
\label{exp:abla}

\subsubsection{Effect of Repulsive Anchors}
As shown in Table~\ref{tab:tab3_component_ablation}, continuing post-training without either proposed component yields 38.2 EPDMS, only 0.5 points above the pretrained baseline. Introducing repulsive anchors increases EPDMS to 39.1, outperforming the matched post-training control and pretrained baseline by 0.9 and 1.4 points, respectively. This result demonstrates that explicitly separating predictions from hard-negative trajectories provides effective safety supervision beyond additional training alone.

\begin{table}[!t]
    \caption{Ablation study of the proposed post-training components.
    A checkmark indicates that the corresponding component is enabled.}
    \label{tab:tab3_component_ablation}
    \centering
    \small
    \setlength{\tabcolsep}{4pt}
    \renewcommand{\arraystretch}{1.15}
    \begin{tabular}{@{}cccc@{}}
        \toprule
        \makecell[cc]{Post-training}
        & \makecell[cc]{Repulsive anchors}
        & \makecell[cc]{Scene-conditioned\\rider tokens}
        & \makecell[cc]{EPDMS} \\
        \midrule
        \multicolumn{3}{c}{Pretrained Baseline} & 37.7 \\
        \midrule
        \cmark &        &        & 38.2 \\
        \cmark & \cmark &        & 39.1   \\
        \cmark &        & \cmark & 38.8   \\
        \cmark & \cmark & \cmark & \textbf{40.4}   \\
        \bottomrule
    \end{tabular}
\end{table}

\subsubsection{Effect of Scene-Conditioned Rider Tokens}
Using scene-conditioned rider tokens alone achieves 38.8 EPDMS, outperforming the post-training control and pretrained baseline by 0.6 and 1.1 points, respectively. Combining rider tokens with repulsive anchors further increases EPDMS to 40.4, exceeding the anchor-only and rider-only variants by 1.3 and 1.6 points, respectively. The additional 1.3 EPDMS gain over the anchor-only variant further shows that NegWAM's improvement cannot be explained by trajectory-space repulsion alone.

\begin{figure}[!t]
    \centering
    \includegraphics[width=\linewidth]{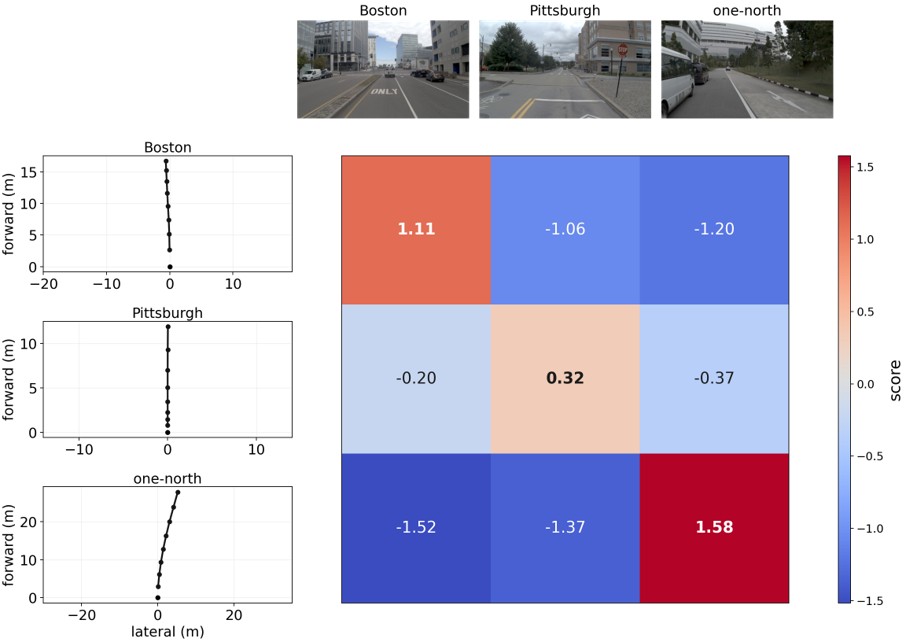}
    \caption{Scene--trajectory compatibility scores for three example scenes. Rows correspond to the trajectories shown on the left, and columns correspond to the observed scenes shown above. Each cell reports the compatibility score of the corresponding scene--trajectory pair.}
    \label{fig:compability}
\end{figure}

Figure~\ref{fig:compability} further illustrates the scene--trajectory compatibility learned by the rider branch. Pairing three trajectories with scenes from Boston, Pittsburgh, and one-north yields the highest score for each trajectory in its corresponding scene: the diagonal scores are 1.11, 0.32, and 1.58, respectively, and all mismatched pairs receive negative scores. This example complements the component ablation by illustrating the scene dependence of the learned compatibility scores.

\subsubsection{Sensitivity to Post-Training Budget and Hyperparameters}
\begin{figure}[!t]
\centering
\includegraphics[width=0.9\linewidth]
{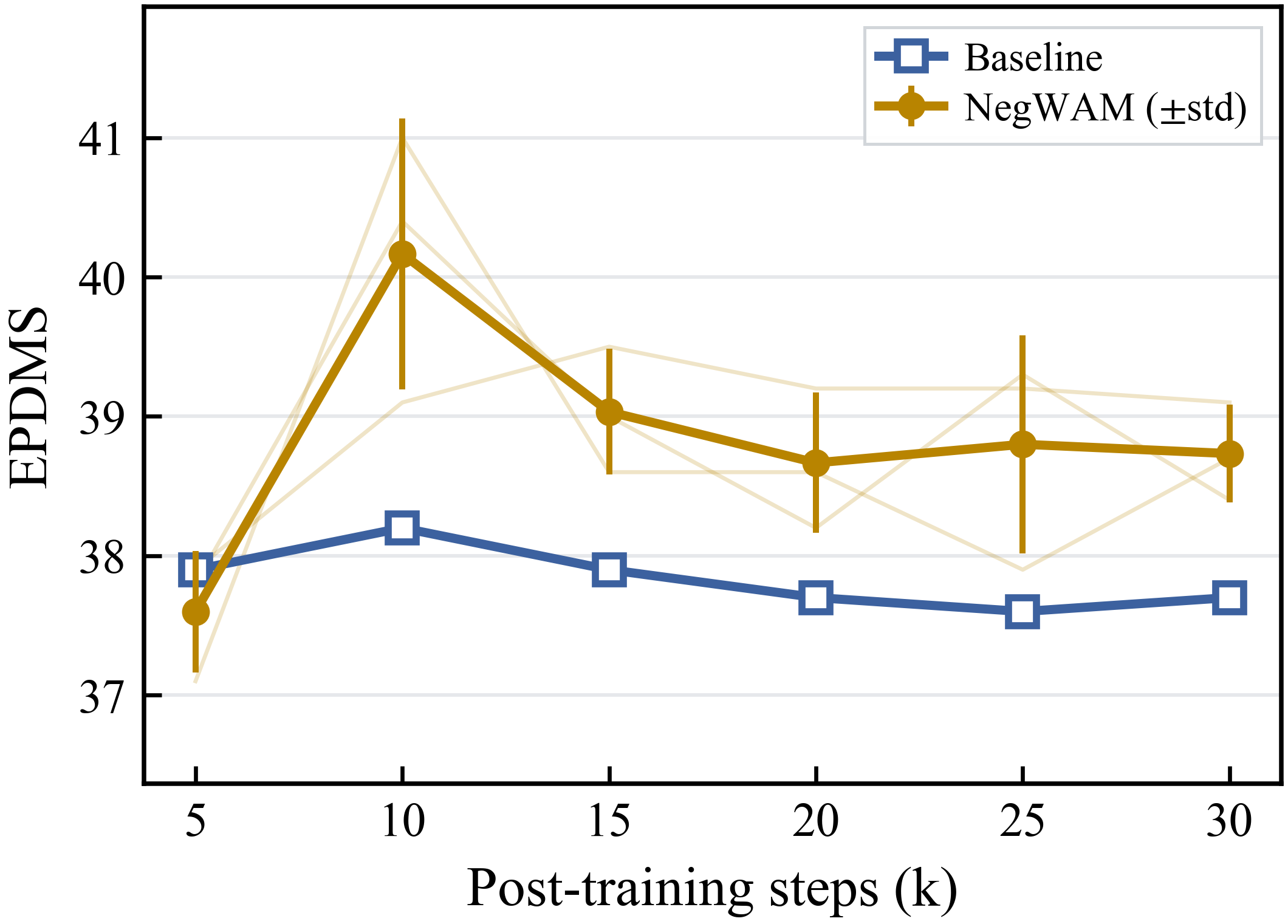}
\caption{
Effect of Post-Training Steps for NAVSIM v2 \textit{navhard}. NegWAM results are reported as mean $\pm$ one standard deviation over three random seeds, while the baseline is a single run.
}
\label{fig:fig4_params}
\end{figure}

\begin{figure}[!t]
\centering
\includegraphics[width=0.9\linewidth]
{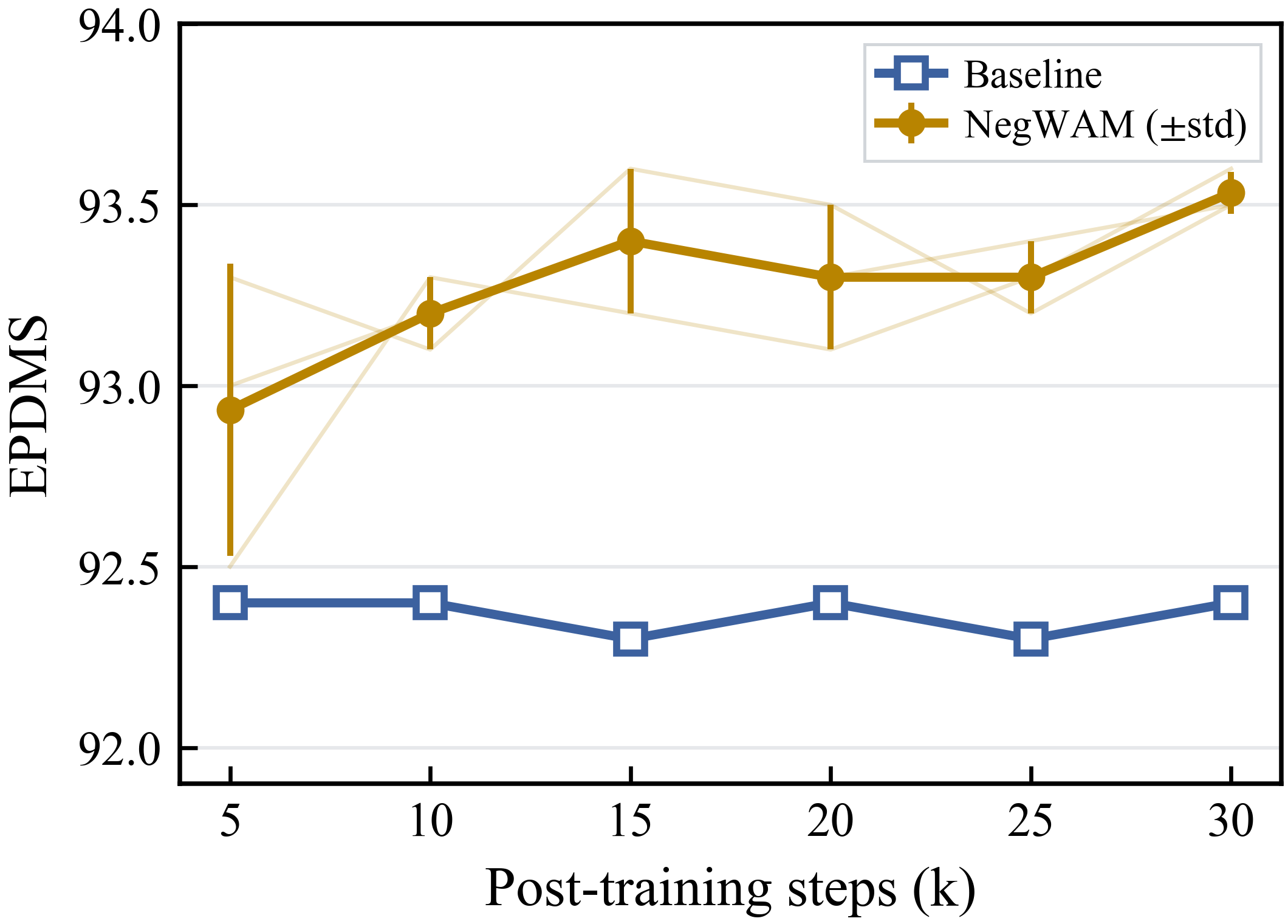}
\caption{
Effect of Post-Training Steps for NAVSIMv2 \textit{navtest}. NegWAM results are reported as mean $\pm$ one standard deviation over three random seeds, while the baseline is a single run.
}
\label{fig:navtest_posttraining_steps}
\end{figure}

Following the public NAVSIM v2 protocol, which does not define a separate validation split, we report the complete checkpoint and one-factor-at-a-time sensitivity results on \textit{navhard} for transparency. As shown in Figure~\ref{fig:fig4_params}, NegWAM consistently outperforms the matched standard post-training baseline from 10K to 30K steps, confirming that the improvement holds across the post-training budget rather than at a single checkpoint. The corresponding \textit{navtest} results in Figure~\ref{fig:navtest_posttraining_steps} also show consistent gains over the baseline from 5K to 30K steps. Table~\ref{tab:loss_weight_sensitivity} varies the learning rate, $\lambda_{\mathrm{rep}}$, and $\lambda_{\mathrm{compat}}$ around the default configuration used in the main experiments. All tested settings yield 39.0--40.4 EPDMS, showing robustness to moderate hyperparameter changes.


\begin{table}[!t]
    \caption{
    Sensitivity to the learning rate, repulsion weight, and compatibility weight.}
    \label{tab:loss_weight_sensitivity}
    \centering
    \small
    \setlength{\tabcolsep}{5pt}
    \renewcommand{\arraystretch}{1.12}
    \begin{tabular}{@{}cccccc@{}}
        \toprule
        \multicolumn{2}{c}{Learning rate}
        & \multicolumn{2}{c}{Repulsion}
        & \multicolumn{2}{c}{Compatibility} \\
        \cmidrule(lr){1-2}
        \cmidrule(lr){3-4}
        \cmidrule(lr){5-6}
        Learning rate
        & EPDMS
        & $\lambda_{\mathrm{rep}}$
        & EPDMS
        & $\lambda_{\mathrm{compat}}$
        & EPDMS \\
        \midrule
        $5{\times}10^{-7}$ & 39.5          
        & 50  & 39.0          
        & 0.05 & 39.3 \\
        
        $1{\times}10^{-6}$ & \textbf{40.4} 
        & 100 & \textbf{40.4} 
        & 0.1  & \textbf{40.4} \\
        
        $2{\times}10^{-6}$ & 39.1          
        & 150 & 39.8          
        & 0.2  & 39.1 \\
        
        \bottomrule
    \end{tabular}
\end{table}

\subsection{Qualitative Analysis}
\label{exp:quali}

We examine NegWAM's behavior through trajectory comparisons and a representative closed-loop rollout.

\subsubsection{Trajectory Comparison}
\begin{figure*}[!t]
\centering
\includegraphics[width=\textwidth]
{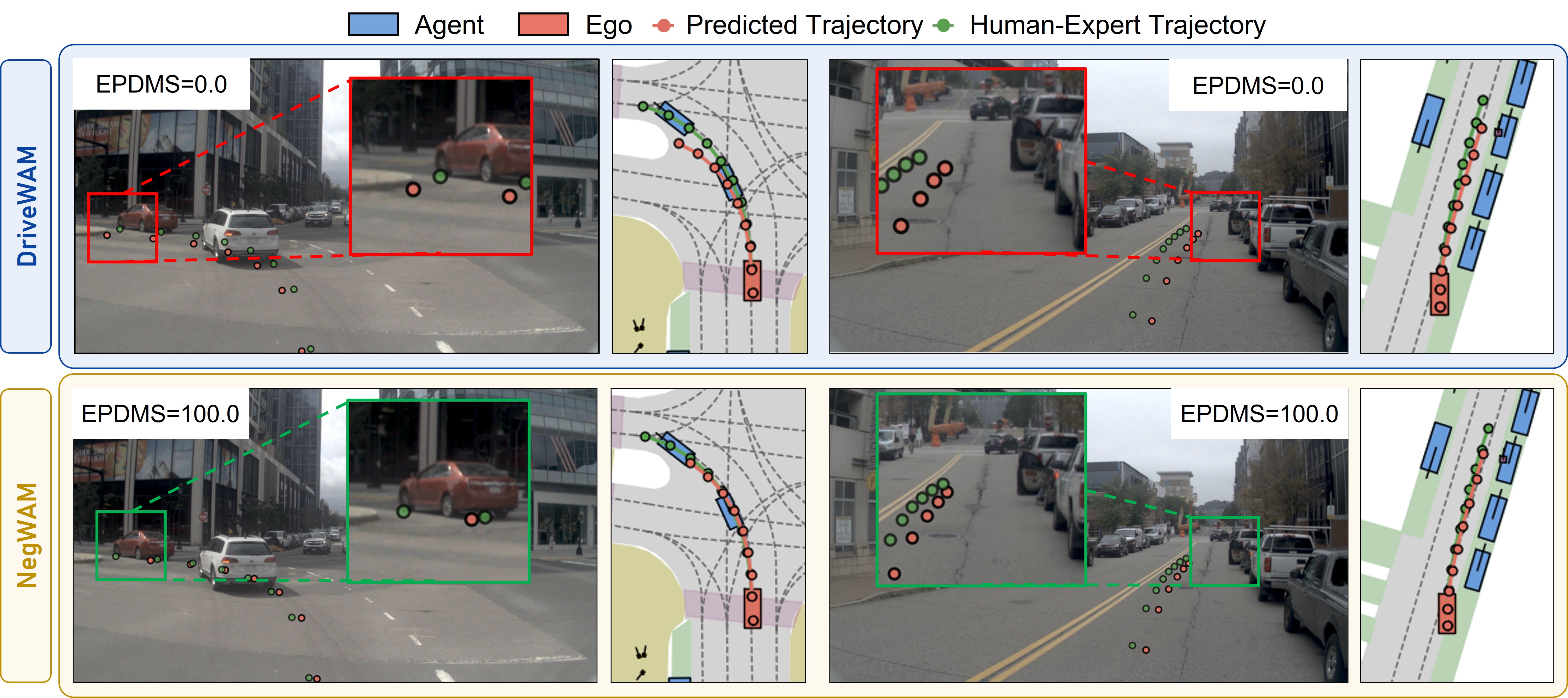}
\caption{
Qualitative comparison between DriveWAM~\cite{shi2026drivewam} and our NegWAM in safety-critical scenarios.
DriveWAM (top) predicts trajectories that approach nearby traffic participants, whereas NegWAM (bottom) maintains safer clearance while better following the human-expert trajectory.
}
\label{fig:fig3_qualita}
\end{figure*}

As shown in Figure~\ref{fig:fig3_qualita}, NegWAM maintains safer clearance than DriveWAM~\cite{shi2026drivewam} and avoids potential collision regions. In the second scenario, an occupant of a roadside parked vehicle opens the door and begins to exit. DriveWAM~\cite{shi2026drivewam} still produces a nominal close-passing trajectory, leaving insufficient clearance for this emerging hazard. After post-training, NegWAM instead shifts outward and maintains a safe distance from the parked vehicle. Although the baseline trajectory may be safe without the door-opening interaction, it entails substantial risk in this safety-critical context. This comparison suggests that NegWAM internalizes scene-dependent safety boundaries that are difficult to capture through expert imitation alone, enabling it to avoid trajectories that are geometrically close to expert behavior yet potentially unsafe.

\subsubsection{Closed-Loop Rollout in HUGSIM}
\begin{figure}[!t]
\centering
\includegraphics[width=\linewidth]{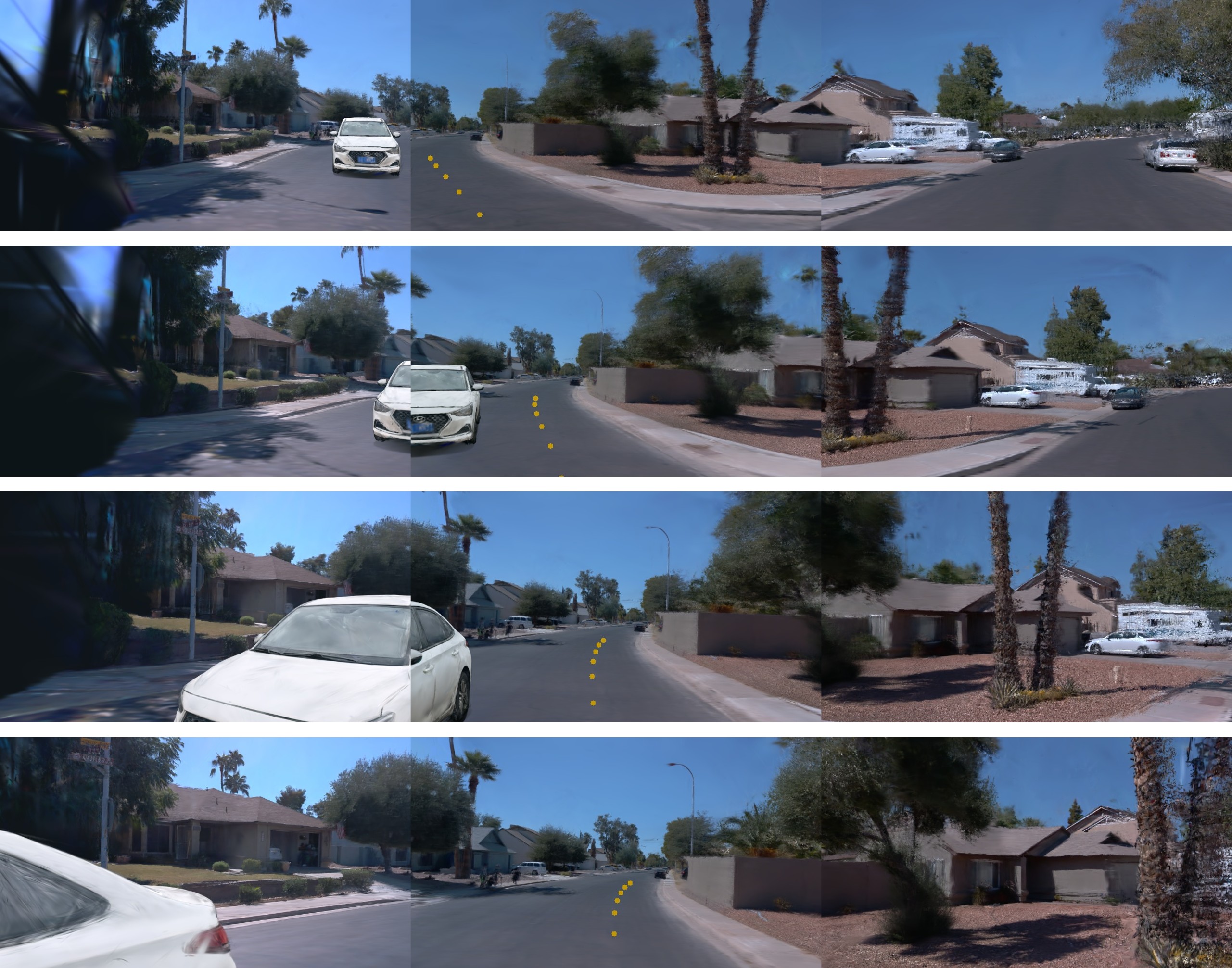}
\caption{Representative closed-loop rollout in HUGSIM. Rows are ordered chronologically from top to bottom, with three rendered camera views at each time point. Yellow dots mark the overlaid trajectory in the center view. An oncoming vehicle approaches and passes on the left as the ego vehicle proceeds along the road.}
\label{fig:hugsim_closed_loop}
\end{figure}

Figure~\ref{fig:hugsim_closed_loop} shows a representative closed-loop rollout in HUGSIM. Across the displayed time points, an oncoming vehicle approaches and passes on the left while the ego vehicle continues along the road. The rendered observations and trajectory overlays illustrate how the driving scene evolves during closed-loop execution.

\begin{table}[!t]
\centering\footnotesize
\caption{Planning score on navhard and inference efficiency.
Higher EPDMS and FPS and lower latency, model size, and peak GPU memory are preferred.}
\label{tab:inference_efficiency}
\setlength{\tabcolsep}{3pt}
\renewcommand{\arraystretch}{1.1}
\resizebox{\linewidth}{!}{%
\begin{tabular}{@{}lrrrrr@{}}
\toprule
Method
& \shortstack{Latency\\(ms) $\downarrow$}
& FPS $\uparrow$
& \shortstack{Model size\\(MB) $\downarrow$}
& \shortstack{Peak GPU\\memory (MB) $\downarrow$}
& EPDMS $\uparrow$ \\
\midrule
DriveWAM~\cite{shi2026drivewam}
& 871.35 & 1.15 & 6241.81 & 14029.08 & 37.7 \\
NegWAM
& 877.86 & 1.14 & 6241.81 & 14029.08 & \textbf{40.4} \\
\bottomrule
\end{tabular}%
}
\end{table}

\subsection{Inference Efficiency}
\label{exp:efficiency}

Table~\ref{tab:inference_efficiency} compares planning performance on \textit{navhard} and inference efficiency. NegWAM improves EPDMS from 37.7 to 40.4 while retaining the same model storage size and peak inference GPU memory usage. Measured latency increases from 871.35 to 877.86\,ms, a difference of 6.51\,ms ($0.75\,\%$), while throughput changes from 1.15 to 1.14 FPS. The auxiliary rider groups and compatibility head are removed at inference, allowing the planning gains to be obtained with nearly unchanged deployment cost.

\section{Conclusion}
\label{sec:conclusion}

We revisit unified world--action models from the perspective of safety post-training and introduce NegWAM to address their reliance on expert imitation and lack of explicit safety boundaries. NegWAM leverages offline-constructed, near-expert hard-negative trajectories as repulsive anchors and scene-conditioned rider tokens, internalizing scene-dependent safety boundaries at both the trajectory and representation levels. Experiments demonstrate that NegWAM improves robustness in challenging scenarios while preserving routine driving performance, without requiring online simulation during post-training or introducing additional inference overhead. Zero-shot closed-loop evaluation further validates the transferability of these learned safety boundaries to unseen interactive environments.

The generalization of NegWAM to the diversity and unpredictability of real-world driving remains to be established. Learning safety boundaries that remain effective across evolving environments and complex traffic interactions is a broader challenge for world--action models. Future work will investigate more generalizable and adaptive safety learning paradigms, toward integrating predictive world understanding with reliable long-horizon decision-making for autonomous driving.

\bibliographystyle{IEEEtran}
\bibliography{reference}

@article{chen2024end-to-end,
  title={End-to-end autonomous driving: Challenges and frontiers},
  author={Chen, Li and Wu, Penghao and Chitta, Kashyap and Jaeger, Bernhard and Geiger, Andreas and Li, Hongyang},
  journal={IEEE Trans. Pattern Anal. Mach. Intell.},
  year={2024},
}

@misc{karkus2025beyond,
  title        = {Beyond Behavior Cloning in Autonomous Driving: a Survey of Closed-Loop Training Techniques},
  author       = {Karkus, Peter and Igl, Maximilian and Chen, Yuxiao and Chitta, Kashyap and Packer, Jef and Douillard, Bertrand and Tian, Ran and Naumann, Alexander and Garcia-Cobo, Guillermo and Tan, Shuhan and others},
  year         = {2025},
  howpublished = {Authorea Preprints},
  publisher    = {Authorea},
}

@misc{yang2026posttraining,
  title         = {Post-Training in End-to-End Autonomous Driving},
  author        = {Yang, Ruining and Wang, Muxing and Chen, Yixiao and Guo, Tongfei and Xu, Yi and Cui, Can and Yang, Zichong and Zhang, Yitian and Wang, Ziran and Fu, Yun and Su, Lili},
  year          = {2026},
  eprint        = {2607.08072},
  archivePrefix = {arXiv},
  url           = {https://arxiv.org/abs/2607.08072}
}

@inproceedings{dosovitskiy2017carla,
  title={CARLA: An open urban driving simulator},
  author={Dosovitskiy, Alexey and Ros, German and Codevilla, Felipe and Lopez, Antonio and Koltun, Vladlen},
  booktitle={Proc. Conf. Robot Learn. (CoRL)},
  pages={1--16},
  year={2017},
  organization={PMLR}
}

@inproceedings{zhang2021end,
  title={End-to-end urban driving by imitating a reinforcement learning coach},
  author={Zhang, Zhejun and Liniger, Alexander and Dai, Dengxin and Yu, Fisher and Van Gool, Luc},
  booktitle={Proc. IEEE/CVF Int. Conf. Comput. Vis. (ICCV)},
  pages={15222--15232},
  year={2021}
}

@inproceedings{gao2026rad,
  title={Rad: Training an end-to-end driving policy via large-scale 3dgs-based reinforcement learning},
  author={Gao, Hao and Chen, Shaoyu and Jiang, Bo and Liao, Bencheng and Shi, Yiang and Guo, Xiaoyang and Pu, Yuechuan and Li, Xiangyu and Liu, Wenyu and Zhang, Qian and others},
  booktitle={Adv. Neural Inf. Process. Syst. (NeurIPS)},
  pages={32551--32576},
  year={2025}
}

@misc{li2026world,
  title         = {World Engine: Towards the Era of Post-Training for Autonomous Driving},
  author        = {Li, Tianyu and Chen, Li and Wang, Caojun and Liu, Haochen and Chitta, Kashyap and Yang, Zhenjie and Lu, Yuhang and Ye, Naisheng and Qiu, Yihang and Wang, Yufei and others},
  year          = {2026},
  eprint        = {2606.19836},
  archivePrefix = {arXiv},
  url           = {https://arxiv.org/abs/2606.19836}
}

@misc{yang2026dreamerad,
  title         = {{DreamerAD}: Efficient Reinforcement Learning via Latent World Model for Autonomous Driving},
  author        = {Yang, Pengxuan and Zheng, Yupeng and Qian, Deheng and Xing, Zebin and Zhang, Qichao and Wang, Linbo and Zhang, Yichen and Guo, Shaoyu and Xia, Zhongpu and Chen, Qiang and others},
  year          = {2026},
  eprint        = {2603.24587},
  archivePrefix = {arXiv},
  url           = {https://arxiv.org/abs/2603.24587}
}

@inproceedings{yang2026worldrft,
  title={Worldrft: Latent world model planning with reinforcement fine-tuning for autonomous driving},
  author={Yang, Pengxuan and Lu, Ben and Xia, Zhongpu and Han, Chao and Gao, Yinfeng and Zhang, Teng and Zhan, Kun and Lang, XianPeng and Zheng, Yupeng and Zhang, Qichao},
  booktitle={Proc. AAAI Conf. Artif. Intell. (AAAI)},
  volume={40},
  number={14},
  pages={11649--11657},
  year={2026}
}

@misc{xu2026toad,
  title         = {Test-Time Trajectory Optimization for Autonomous Driving},
  author        = {Xu, Yihong and Zablocki, Eloi and Yin, Yuan and Ramzi, Elias and Kirby, Ellington and Boulch, Alexandre and Cord, Matthieu},
  year          = {2026},
  eprint        = {2606.07170},
  archivePrefix = {arXiv},
  url           = {https://arxiv.org/abs/2606.07170}
}

@misc{wang2026beyond,
  title         = {Beyond Imitation: Learning Safe End-to-End Autonomous Driving from Hard Negatives},
  author        = {Wang, Junli and Hua, Zhihua and Liu, Xueyi and Xing, Zebin and Tian, Haochen and Ma, Kun and Ye, Hangjun and Chen, Guang and Chen, Long and Zhang, Qichao},
  year          = {2026},
  eprint        = {2605.19771},
  archivePrefix = {arXiv},
  url           = {https://arxiv.org/abs/2605.19771}
}

@inproceedings{shang2026drivedpo,
  title={Drivedpo: Policy learning via safety dpo for end-to-end autonomous driving},
  author={Shang, Shuyao and Chen, Yuntao and Wang, Yuqi and Li, Yingyan and ZHANG, ZHAO-XIANG},
  booktitle={Adv. Neural Inf. Process. Syst. (NeurIPS)},
  pages={81565--81585},
  year={2025}
}

@article{liu2025takead,
  title={TakeAD: Preference-Based Post-Optimization for End-to-End Autonomous Driving With Expert Takeover Data},
  author={Liu, Deqing and Gao, Yinfeng and Qian, Deheng and Zhang, Qichao and Ye, Xiaoqing and Han, Junyu and Zheng, Yupeng and Liu, Xueyi and Xia, Zhongpu and Ding, Dawei and others},
  journal={IEEE Robot. Autom. Lett.},
  volume={11},
  number={2},
  pages={1738--1745},
  year={2025},
  publisher={IEEE}
}

@inproceedings{hu2023uniad,
  title={Planning-oriented autonomous driving},
  author={Hu, Yihan and Yang, Jiazhi and Chen, Li and Li, Keyu and Sima, Chonghao and Zhu, Xizhou and Chai, Siqi and Du, Senyao and Lin, Tianwei and Wang, Wenhai and others},
  booktitle={Proc. IEEE/CVF Conf. Comput. Vis. Pattern Recognit. (CVPR)},
  pages={17853--17862},
  year={2023}
}

@inproceedings{naumann2025data,
  title={Data scaling laws for end-to-end autonomous driving},
  author={Naumann, Alexander and Gu, Xunjiang and Dimlioglu, Tolga and Bojarski, Mariusz and Degirmenci, Alperen and Popov, Alexander and Bisla, Devansh and Pavone, Marco and Muller, Urs and Ivanovic, Boris},
  booktitle={Proc. IEEE/CVF Conf. Comput. Vis. Pattern Recognit. (CVPR)},
  pages={2571--2582},
  year={2025}
}

@inproceedings{zhang2025epona,
  title={Epona: Autoregressive diffusion world model for autonomous driving},
  author={Zhang, Kaiwen and Tang, Zhenyu and Hu, Xiaotao and Pan, Xingang and Guo, Xiaoyang and Liu, Yuan and Huang, Jingwei and Yuan, Li and Zhang, Qian and Long, Xiao-Xiao and others},
  booktitle={Proc. IEEE/CVF Int. Conf. Comput. Vis. (ICCV)},
  pages={27220--27230},
  year={2025}
}

@inproceedings{wang2024drivedreamer,
  title={Drivedreamer: Towards real-world-drive world models for autonomous driving},
  author={Wang, Xiaofeng and Zhu, Zheng and Huang, Guan and Chen, Xinze and Zhu, Jiagang and Lu, Jiwen},
  booktitle={Proc. Eur. Conf. Comput. Vis. (ECCV)},
  pages={55--72},
  year={2024},
  organization={Springer}
}

@inproceedings{yang2026geniedrive,
  title={Geniedrive: Towards physics-aware driving world model with 4d occupancy guided video generation},
  author={Yang, Zhenya and Liu, Zhe and Lu, Yuxiang and Hou, Liping and Miao, Chenxuan and Peng, Siyi and Feng, Bailan and Bai, Xiang and Zhao, Hengshuang},
  booktitle={Proc. IEEE/CVF Conf. Comput. Vis. Pattern Recognit. (CVPR)},
  pages={35680--35690},
  year={2026}
}

@inproceedings{yang2026resim,
  title     = {{ReSim}: Reliable World Simulation for Autonomous Driving},
  author    = {Yang, Jiazhi and Chitta, Kashyap and Gao, Shenyuan and Chen, Long and Shao, Yuqian and Jia, Xiaosong and Li, Hongyang and Geiger, Andreas and Yue, Xiangyu and Chen, Li},
  booktitle = {Adv. Neural Inf. Process. Syst. (NeurIPS)},
  pages     = {167710--167741},
  year      = {2025}
}

@inproceedings{zhou2024dino,
  title     = {{DINO-WM}: World Models on Pre-trained Visual Features Enable Zero-Shot Planning},
  author    = {Zhou, Gaoyue and Pan, Hengkai and LeCun, Yann and Pinto, Lerrel},
  booktitle = {Proc. 42nd Int. Conf. Mach. Learn. (ICML)},
  pages     = {79115--79135},
  year      = {2025}
}

@misc{wang2026latent,
  title         = {{Latent-WAM}: Latent World Action Modeling for End-to-End Autonomous Driving},
  author        = {Wang, Linbo and Zheng, Yupeng and Chen, Qiang and Li, Shiwei and Zhang, Yichen and Xing, Zebin and Zhang, Qichao and Li, Xiang and Qian, Deheng and Yang, Pengxuan and others},
  year          = {2026},
  eprint        = {2603.24581},
  archivePrefix = {arXiv},
  url           = {https://arxiv.org/abs/2603.24581}
}

@misc{wang2026drive,
  title         = {{Drive-JEPA}: Video {JEPA} Meets Multimodal Trajectory Distillation for End-to-End Driving},
  author        = {Wang, Linhan and Yang, Zichong and Bai, Chen and Zhang, Guoxiang and Liu, Xiaotong and Zheng, Xiaoyin and Long, Xiao-Xiao and Lu, Chang-Tien and Lu, Cheng},
  year          = {2026},
  eprint        = {2601.22032},
  archivePrefix = {arXiv},
  url           = {https://arxiv.org/abs/2601.22032}
}

@misc{zhou2026drivedreamer,
  title         = {{DriveDreamer-Policy}: A Geometry-Grounded World-Action Model for Unified Generation and Planning},
  author        = {Zhou, Yang and Wang, Xiaofeng and Shao, Hao and Wang, Letian and Zhao, Guosheng and Shao, Jiangnan and Zhu, Jiagang and Yu, Tingdong and Zhu, Zheng and Huang, Guan and others},
  year          = {2026},
  eprint        = {2604.01765},
  archivePrefix = {arXiv},
  url           = {https://arxiv.org/abs/2604.01765}
}

@misc{xu2026eponav2,
  title         = {{EponaV2}: Driving World Model with Comprehensive Future Reasoning},
  author        = {Xu, Jiawei and Zhong, Zhizhou and Shu, Zhijian and Jia, Mingkai and Li, Mingxiao and Bian, Jia-Wang and Zhang, Qian and Zhang, Kaicheng and Xie, Jin and Yang, Jian and others},
  year          = {2026},
  eprint        = {2605.14696},
  archivePrefix = {arXiv},
  url           = {https://arxiv.org/abs/2605.14696}
}

@inproceedings{lipman2022flow,
  title     = {Flow Matching for Generative Modeling},
  author    = {Lipman, Yaron and Chen, Ricky T. Q. and Ben-Hamu, Heli and Nickel, Maximilian and Le, Matt},
  booktitle = {Proc. Int. Conf. Learn. Represent. (ICLR)},
  year      = {2023}
}

@inproceedings{liu2022flow,
  title     = {Flow Straight and Fast: Learning to Generate and Transfer Data with Rectified Flow},
  author    = {Liu, Xingchao and Gong, Chengyue and Liu, Qiang},
  booktitle = {Proc. Int. Conf. Learn. Represent. (ICLR)},
  year      = {2023}
}

@article{tong2023improving,
  title   = {Improving and Generalizing Flow-Based Generative Models with Minibatch Optimal Transport},
  author  = {Tong, Alexander and Fatras, Kilian and Malkin, Nikolay and Huguet, Guillaume and Zhang, Yanlei and Rector-Brooks, Jarrid and Wolf, Guy and Bengio, Yoshua},
  journal = {Trans. Mach. Learn. Res.},
  year    = {2024},
}

@misc{li2026causal,
  title         = {Causal World Modeling for Robot Control},
  author        = {Li, Lin and Zhang, Qihang and Luo, Yiming and Yang, Shuai and Wang, Ruilin and Han, Fei and Yu, Mingrui and Gao, Zelin and Xue, Nan and Zhu, Xing and others},
  year          = {2026},
  eprint        = {2601.21998},
  archivePrefix = {arXiv},
  url           = {https://arxiv.org/abs/2601.21998}
}

@misc{li2025gtrs,
  title         = {Generalized Trajectory Scoring for End-to-End Multimodal Planning},
  author        = {Li, Zhenxin and Yao, Wenhao and Wang, Zi and Sun, Xinglong and Chen, Joshua and Chang, Nadine and Shen, Maying and Wu, Zuxuan and Lan, Shiyi and Alvarez, Jose M.},
  year          = {2025},
  eprint        = {2506.06664},
  archivePrefix = {arXiv},
  url           = {https://arxiv.org/abs/2506.06664}
}

@inproceedings{liu2026driveva,
  title = {{DriveVA}: Video action models are zero-shot drivers},
  author = {Liu, Mengmeng and Zhang, Diankun and Liu, Jiuming and Cui, Jianfeng and Xie, Hongwei and Chen, Guang and Ye, Hangjun and Yang, Michael Ying and Nex, Francesco and Cheng, Hao},
  booktitle = {Proc. Eur. Conf. Comput. Vis. (ECCV)},
  pages = {315--335},
  year = {2026},
  doi = {10.1007/978-3-032-37718-0_19}
}

@inproceedings{rafailov2023dpo,
  title     = {Direct Preference Optimization: Your Language Model Is Secretly a Reward Model},
  author    = {Rafailov, Rafael and Sharma, Archit and Mitchell, Eric and Manning, Christopher D. and Ermon, Stefano and Finn, Chelsea},
  booktitle = {Adv. Neural Inf. Process. Syst. (NeurIPS)},
  volume    = {36},
  pages     = {53728--53741},
  year      = {2023}
}

@article{kerbl20233dgs,
  title={3d gaussian splatting for real-time radiance field rendering.},
  author={Kerbl, Bernhard and Kopanas, Georgios and Leimk{\"u}hler, Thomas and Drettakis, George and others},
  journal={ACM Trans. Graph.},
  volume={42},
  number={4},
  pages={139--1},
  year={2023}
}

@misc{gao2026learning,
  title         = {Learning from Mistakes: Post-Training for Driving {VLA} with Takeover Data},
  author        = {Gao, Yinfeng and Liu, Deqing and Zhang, Qichao and Zheng, Yupeng and Tian, Haochen and Li, Guang and Ye, Hangjun and Chen, Long and Ding, Da-Wei and Zhao, Dongbin},
  year          = {2026},
  eprint        = {2603.14972},
  archivePrefix = {arXiv},
  url           = {https://arxiv.org/abs/2603.14972}
}

@misc{hong2026drivefuture,
  title         = {{DriveFuture}: Future-Aware Latent World Models for Autonomous Driving},
  author        = {Hong, Yufeng and Zhou, Xiaotian and Li, Yingyan and Zhou, Xiangpo and Liu, Lin and Luo, Yadan and Xu, Shaoqing and Yang, Lei and Song, Ziying},
  year          = {2026},
  eprint        = {2605.09701},
  archivePrefix = {arXiv},
  url           = {https://arxiv.org/abs/2605.09701}
}

@inproceedings{li2025recogdrive,
  title     = {{ReCogDrive}: A Reinforced Cognitive Framework for End-to-End Autonomous Driving},
  author    = {Li, Yongkang and Xiong, Kaixin and Guo, Xiangyu and Li, Fang and Yan, Sixu and Xu, Gangwei and Zhou, Lijun and Chen, Long and Sun, Haiyang and Wang, Bing and others},
  booktitle = {Proc. Int. Conf. Learn. Represent. (ICLR)},
  year      = {2026}
}

@misc{liu2026driveworld,
  title         = {{DriveWorld-VLA}: Unified Latent-Space World Modeling with Vision-Language-Action for Autonomous Driving},
  author        = {Liu, Lin and Song, Ziying and Jia, Caiyan and Ye, Hangjun and Hao, Xiaoshuai and Chen, Long and others},
  year          = {2026},
  eprint        = {2602.06521},
  archivePrefix = {arXiv},
  url           = {https://arxiv.org/abs/2602.06521}
}

@inproceedings{liao2025diffusiondrive,
  title={Diffusiondrive: Truncated diffusion model for end-to-end autonomous driving},
  author={Liao, Bencheng and Chen, Shaoyu and Yin, Haoran and Jiang, Bo and Wang, Cheng and Yan, Sixu and Zhang, Xinbang and Li, Xiangyu and Zhang, Ying and Zhang, Qian and others},
  booktitle={Proc. IEEE/CVF Conf. Comput. Vis. Pattern Recognit. (CVPR)},
  pages={12037--12047},
  year={2025}
}

@misc{zou2025diffusiondrivev2,
  title         = {{DiffusionDriveV2}: Reinforcement Learning-Constrained Truncated Diffusion Modeling in End-to-End Autonomous Driving},
  author        = {Zou, Jialv and Chen, Shaoyu and Liao, Bencheng and Zheng, Zhiyu and Song, Yuehao and Zhang, Lefei and Zhang, Qian and Liu, Wenyu and Wang, Xinggang},
  year          = {2025},
  eprint        = {2512.07745},
  archivePrefix = {arXiv},
  url           = {https://arxiv.org/abs/2512.07745}
}

@article{yao2026discrete,
  title={Discrete-WAM: Unified Discrete Vision-Action Token Editing for World-Policy Learning},
  author={Yao, Ziyang and Liu, Haochen and Jiang, Yuncheng and Zhu, Zeyu and Guo, Zibin and Wang, Jingru and Liu, Tianle and Cui, Jianwei and Yang, Kuiyuan and Xie, Hongwei and others},
  journal={arXiv preprint arXiv:2606.05645},
  year={2026}
}

@inproceedings{li2025drivevla,
  title     = {{DriveVLA-W0}: World Models Amplify Data Scaling Law in Autonomous Driving},
  author    = {Li, Yingyan and Shang, Shuyao and Liu, Weisong and Zhan, Bing and Wang, Haochen and Wang, Yuqi and Chen, Yuntao and Wang, Xiaoman and An, Yasong and Tang, Chufeng and others},
  booktitle = {Proc. Int. Conf. Learn. Represent. (ICLR)},
  year      = {2026}
}

@inproceedings{liu2026guideflow,
  title={Guideflow: Constraint-guided flow matching for planning in end-to-end autonomous driving},
  author={Liu, Lin and Jia, Caiyan and Yu, Guanyi and Song, Ziying and Li, JunQiao and Jia, Feiyang and Wu, Peiliang and Hao, Xiaoshuai and Luo, Yadan},
  booktitle={Proc. IEEE/CVF Conf. Comput. Vis. Pattern Recognit. (CVPR)},
  pages={3719--3728},
  year={2026}
}

@misc{zhou2026spanvla,
  title         = {{SpanVLA}: Efficient Action Bridging and Learning from Negative-Recovery Samples for Vision-Language-Action Model},
  author        = {Zhou, Zewei and Yang, Ruining and Guo, Yiluan and Chen, Sherry X. and Feng, Tao and Pistunova, Kateryna and Shen, Yishan and Su, Lili and Ma, Jiaqi and others},
  year          = {2026},
  eprint        = {2604.19710},
  archivePrefix = {arXiv},
  url           = {https://arxiv.org/abs/2604.19710}
}

@misc{wan2025wan,
  title         = {Wan: Open and Advanced Large-Scale Video Generative Models},
  author        = {Wan, Team and Wang, Ang and Ai, Baole and Wen, Bin and Mao, Chaojie and Xie, Chen-Wei and Chen, Di and Yu, Feiwu and Zhao, Haiming and Yang, Jianxiao and others},
  year          = {2025},
  eprint        = {2503.20314},
  archivePrefix = {arXiv},
  url           = {https://arxiv.org/abs/2503.20314}
}

@inproceedings{jiang2023vad,
  title={Vad: Vectorized scene representation for efficient autonomous driving},
  author={Jiang, Bo and Chen, Shaoyu and Xu, Qing and Liao, Bencheng and Chen, Jiajie and Zhou, Helong and Zhang, Qian and Liu, Wenyu and Huang, Chang and Wang, Xinggang},
  booktitle={Proc. IEEE/CVF Int. Conf. Comput. Vis. (ICCV)},
  pages={8340--8350},
  year={2023}
}

@inproceedings{cao2025navsimv2,
  title={Pseudosimulation for autonomous driving},
  author={Cao, Wei and Hallgarten, Marcel and Li, Tianyu and Dauner, Daniel and Gu, Xunjiang and Wang, Caojun and Miron, Yakov and Aiello, Marco and Li, Hongyang and Gilitschenski, Igor and Ivanovic, Boris and Pavone, Marco and Geiger, Andreas and Chitta, Kashyap},
  booktitle={Proc. Conf. Robot Learn. (CoRL)},
  year={2025}
}

@inproceedings{dauner2024navsim,
  title={Navsim: Data-driven nonreactive autonomous vehicle simulation and benchmarking},
  author={Dauner, Daniel and Hallgarten, Marcel and Li, Tianyu and Weng, Xinshuo and Huang, Zhiyu and Yang, Zetong and Li, Hongyang and Gilitschenski, Igor and Ivanovic, Boris and Pavone, Marco and Geiger, Andreas and Chitta, Kashyap},
  booktitle={Adv. Neural Inf. Process. Syst. (NeurIPS)},
  year={2024}
}

@misc{openscene2023,
  title        = {OpenScene: The largest up-to-date 3d occupancy prediction benchmark in autonomous driving},
  author       = {{OpenScene Contributors}},
  howpublished = {\url{https://github.com/OpenDriveLab/OpenScene}},
  year         = {2023}
}

@inproceedings{karnchanachari2024towards,
  title={Towards learning-based planning: The nuplan benchmark for real-world autonomous driving},
  author={Karnchanachari, Napat and Geromichalos, Dimitris and Tan, Kok Seang and Li, Nanxiang and Eriksen, Christopher and Yaghoubi, Shakiba and Mehdipour, Noushin and Bernasconi, Gianmarco and Fong, Whye Kit and Guo, Yiluan and others},
  booktitle={Proc. IEEE Int. Conf. Robot. Autom. (ICRA)},
  year={2024}
}

@misc{li2025hydramdpplusplus,
  title         = {{Hydra-MDP++}: Advancing End-to-End Driving via Expert-Guided Hydra-Distillation},
  author        = {Li, Kailin and Li, Zhenxin and Lan, Shiyi and Xie, Yuan and Zhang, Zhizhong and Liu, Jiayi and Wu, Zuxuan and Yu, Zhiding and Alvarez, Jose M.},
  year          = {2025},
  eprint        = {2503.12820},
  archivePrefix = {arXiv},
  url           = {https://arxiv.org/abs/2503.12820}
}

@inproceedings{xia2026drivelaw,
  title={Drivelaw: Unifying planning and video generation in a latent driving world},
  author={Xia, Tianze and Li, Yongkang and Zhou, Lijun and Yao, Jingfeng and Xiong, Kaixin and Sun, Haiyang and Wang, Bing and Ma, Kun and Chen, Guang and Ye, Hangjun and others},
  booktitle={Proc. IEEE/CVF Conf. Comput. Vis. Pattern Recognit. (CVPR)},
  pages={39701--39712},
  year={2026}
}

@inproceedings{shi2026drivewam,
  title = {{DriveWAM}: Video generative priors enable scalable world-action modeling for autonomous driving},
  author = {Shi, Chen and Xu, Jinrui and Shi, Shaoshuai and Sheng, Kehua and Zhang, Bo and Jiang, Li},
  booktitle = {Adv. Neural Inf. Process. Syst. (NeurIPS)},
  year = {2026}
}

@inproceedings{karnchanachari2024nuplan,
  title={Towards learning-based planning: The nuplan benchmark for real-world autonomous driving},
  author={Karnchanachari, Napat and Geromichalos, Dimitris and Tan, Kok Seang and Li, Nanxiang and Eriksen, Christopher and Yaghoubi, Shakiba and Mehdipour, Noushin and Bernasconi, Gianmarco and Fong, Whye Kit and Guo, Yiluan and others},
  booktitle={Proc. IEEE Int. Conf. Robot. Autom. (ICRA)},
  year={2024}
}

@article{zhou2025hugsim,
  title={Hugsim: A real-time, photo-realistic and closed-loop simulator for autonomous driving},
  author={Zhou, Hongyu and Lin, Longzhong and Wang, Jiabao and Lu, Yichong and Bai, Dongfeng and Liu, Bingbing and Wang, Yue and Geiger, Andreas and Liao, Yiyi},
  journal={IEEE Trans. Pattern Anal. Mach. Intell. (TPAMI)},
  year={2025},
  publisher={IEEE}
}

@misc{zhang2026vira,
  title = {Do Better Visual Representations Always Lead to Better End-to-End Autonomous Driving?},
  author = {Zhang, Zihao and Tian, Haochen and Li, Tianyu and Jing, Changhui and He, Jingliang and Ye, Naisheng and Pu, Ziyuan and Yang, Zhenjie},
  year = {2026},
  eprint = {2610.09695},
  archivePrefix = {arXiv},
  url = {https://arxiv.org/abs/2610.09695}
}

\appendices
\section{}

This appendix provides temporal indexing and optimization details, additional NAVSIM component scores, and qualitative cases from both trajectory prediction and closed-loop simulation. It also documents the scope of the negative supervision and the resources used in our implementation.

\subsection{Implementation Details}
\label{sec:supp_imple}

\subsubsection{Data Layout and Optimization}
\label{sec:supp_imple_model}

Table~\ref{tab:tables2_model_training_hparams} summarizes the model configuration, optimization settings, and resource requirements of the two training stages. The four future visual targets use zero-based indices $\{1,3,5,7\}$ within the future-frame sequence, corresponding to cache indices $\{2,4,6,8\}$ when the current image occupies index zero. Each action channel $(\Delta x,\Delta y,\Delta\theta)$ is normalized using its 1st and 99th percentiles.

\begin{table*}[t]
    \caption{Model and training configurations.}
    \label{tab:tables2_model_training_hparams}
    \centering
    \resizebox{\textwidth}{!}{
    \begin{tabular}{cccccccc}
        \toprule
        \multicolumn{8}{c}{\textbf{Model Configuration}} \\
        \midrule
        \textbf{Sensors}
        & \textbf{Observed Input}
        & \textbf{Future Indices}
        & \textbf{Action Target}
        & \textbf{SNR Shift (V/A)}
        & \textbf{Denoising Steps (V/A)}
        & \textbf{Model Size}
        & \textbf{Peak GPU Memory} \\
        \midrule
        1 $\times$ Cam. (SV)
        & \makecell{$1\times256\times448$\\4 historical ego poses}
        & $\{1,3,5,7\}$
        & 8-step $\Delta$-pose
        & $5.0\,/\,1.0$
        & $3\,/\,10$
        & 6,241.81 MB
        & 14,029.08 MB \\
        \bottomrule
    \end{tabular}}

    \vspace{4pt}

    \resizebox{\textwidth}{!}{
    \begin{tabular}{lcccccccc}
        \toprule
        \multicolumn{9}{c}{\textbf{Training Configuration}} \\
        \midrule
        \textbf{Stage}
        & \textbf{Initialization}
        & \textbf{Steps}
        & \textbf{Initial LR}
        & \textbf{LR Schedule}
        & \textbf{Loss Weights}
        & \textbf{GPUs}
        & \textbf{Wall Time}
        & \textbf{GPU Hours} \\
        \midrule
        WAM Pretraining
        & Wan2.2-TI2V-5B
        & 100K
        & $1\times10^{-5}$
        & $\times0.5$ at 30K, 70K, 90K
        & --
        & $4\times$ H100
        & 22 h
        & 88 \\

        Safety Post-training
        & Pretrained WAM
        & 10K
        & $1\times10^{-6}$
        & $\times0.5$ at 5K
        & $\lambda_{\mathrm{rep}}=100,\;
           \lambda_{\mathrm{compat}}=0.1$
        & $4\times$ H100
        & 2 h
        & 8 \\
        \bottomrule
    \end{tabular}}

    \vspace{2pt}
    \parbox{\textwidth}{\footnotesize
    Both stages are trained on \textit{navtrain} using AdamW with
    $(\beta_1,\beta_2)=(0.9,0.95)$ and a weight decay of 0.1.
    The per-GPU and global batch sizes are 1 and 4, respectively.
    Hard-negative defaults are $\alpha=0.9$ and $\tau_s=0.1$;
    the compatibility head has a hidden width of 256.}
\end{table*}

\subsubsection{Scheduler and Conditioning}
\label{sec:supp_scheduler}

For each predicted temporal chunk, we uniformly sample an index $j\in\{0,\ldots,S-1\}$ with $S=1000$ and use the shifted noise scale
\begin{equation}
    u_j=1-\frac{j}{S},\qquad
    t_j^m=\frac{\gamma_m u_j}{1+(\gamma_m-1)u_j},
    \label{eq:supp_shifted_noise}
\end{equation}
where $\gamma_v=5.0$ and $\gamma_a=1.0$. Uniform index sampling therefore does not imply a uniform distribution over the shifted video noise scale. Let $g(t)=\exp[-2(t-0.5)^2]$ and $g_{m,\min}=\min_{0\leq l<S}g(t_l^m)$. The flow-matching weight is
\begin{equation}
    w_m(t_j^m)=
    \frac{g(t_j^m)-g_{m,\min}}
    {S^{-1}\sum_{l=0}^{S-1}[g(t_l^m)-g_{m,\min}]},
    \label{eq:supp_fm_weight}
\end{equation}
which has mean one over the scheduler grid. The video conditioning stream receives noise augmentation with probability $0.5$, while its observed prefix remains clean. The action conditioning stream is not augmented.

\subsubsection{Negative-Cache Indexing}
\label{sec:supp_hard_negative_construction}

The offline negative pools preserve the original 40 ego-frame poses of each GTRS candidate~\cite{li2025gtrs}, at $0.1\,\mathrm{s}$ intervals. Let $\bar{\mathbf{p}}_q$ denote the pose at zero-based cache index $q$. The eight-step training trajectory uses
\begin{equation}
    \mathbf{p}_j=\bar{\mathbf{p}}_{5j+4},
    \qquad j=0,\ldots,7,
    \label{eq:supp_cache_alignment}
\end{equation}
corresponding to indices $4,9,\ldots,39$. This aligns negative trajectories with the expert's $0.5\,\mathrm{s}$ prediction grid while retaining the denser representation for offline evaluation. Fixed-size sampling and padding are applied by the training loader rather than stored in the offline pools; the filtering and validity-mask rules are specified in the main algorithm.

\subsection{Additional Quantitative Results}
\label{sec:supp_results_ext}

Table~\ref{tab:tables1_navtest} provides the full component-wise
\textit{navtest} comparison, complementing the condensed comparison in the main paper.

\begin{table*}[!ht]
    \caption{Performance comparison on the NAVSIM v2 \textit{navtest} benchmark.
    All results are reported using the corrected official NAVSIM v2.2 evaluation protocol.
    MV and SV denote multi-view and single-view camera inputs, respectively.
    All evaluated methods use camera-only inputs without LiDAR.}
	\label{tab:tables1_navtest}
	\resizebox{\linewidth}{!}{
		\begin{tabular}{l|c|cccccccccc}
			\toprule
			\textbf{Methods}  
			&{\textbf{Sensor}}
            &{\textbf{NC} $\uparrow$}
            &{\textbf{DAC} $\uparrow$}
            &{\textbf{DDC} $\uparrow$}
            &{\textbf{TLC} $\uparrow$}
            &{\textbf{EP} $\uparrow$}
            &{\textbf{TTC} $\uparrow$}
            &{\textbf{LK} $\uparrow$}
            &{\textbf{HC} $\uparrow$}
            &{\textbf{EC} $\uparrow$}
			& \textbf{EPDMS} $\uparrow$ \\ 

            \midrule
            Human agent
            & {-}
            & {100.0}
            & {100.0}
            & {99.8}
            & {100.0}
            & {97.8}
            & {100.0}
            & {100.0}
            & {98.1}
            & {90.1}
            & {97.9} \\
            \hline
			\multicolumn{12}{l}{\cellcolor[HTML]{DCEAF7} \textit{\textbf{E2E-based Methods}}} 
			\\
            
            DiffusionDrive~\cite{liao2025diffusiondrive}
            & {MV}
            & {98.2}
            & {95.9}
            & {99.4}
            & {99.8}
            & {87.5}
            & {97.3}
            & {96.8}
            & {98.3}
            & {87.7}
            & {84.5} \\

            DiffusionDriveV2~\cite{zou2025diffusiondrivev2}
            & {MV}
            & {97.7}
            & {96.6}
            & {99.2}
            & {99.8}
            & {88.9}
            & {97.2}
            & {96.0}
            & {97.8}
            & {91.0}
            & {87.5} \\

            \hline
			\multicolumn{12}{l}{\cellcolor[HTML]{DCEAF7} \textit{\textbf{VLA-based Methods}}} 
			\\
            Recogdrive~\cite{li2025recogdrive}
            & {MV}
            & {98.3}
            & {95.2}
            & {98.3}
            & {99.8}
            & {87.1}
            & {97.5}
            & {96.6}
            & {99.5}
            & {86.5}
            & {83.6} \\

            DriveVLA-W0~\cite{li2025drivevla}
            & {SV}
            & {98.5}
            & {99.1}
            & {98.0}
            & {99.7}
            & {86.4}
            & {98.1}
            & {93.2}
            & {97.9}
            & {58.9}
            & {86.1} \\

            DriveWorld-VLA~\cite{liu2026driveworld}
            & {MV}
            & {98.6}
            & {99.1}
            & {99.6}
            & {99.8}
            & {87.4}
            & {97.9}
            & {97.0}
            & {97.8}
            & {78.6}
            & {86.8} \\

            \hline
			\multicolumn{12}{l}{\cellcolor[HTML]{DCEAF7} \textit{\textbf{World-Model-based Methods}}} 
			\\
            Latent-WAM~\cite{wang2026latent}
            & \multicolumn{1}{c|}{MV}
            & {98.1}
            & {97.3}
            & {99.6}
            & {99.8}
            & {87.7}
            & {97.3}
            & {97.6}
            & {98.1}
            & {87.3}
            & {89.3} \\

            DriveFuture~\cite{hong2026drivefuture}
            & \multicolumn{1}{c|}{MV}
            & {98.8}
            & {99.1}
            & {99.6}
            & {99.9}
            & {86.6}
            & {98.4}
            & {96.4}
            & {98.3}
            & {74.8}
            & {89.9} \\
            
			Drivelaw~\cite{xia2026drivelaw}
			& \multicolumn{1}{c|}{SV}
            & {98.8}
            & {96.9}
            & {99.6}
            & {99.8}
            & {87.4}
            & {98.3}
            & {97.7}
            & {98.4}
            & {80.1}
			& {88.9}  \\ 

            Discrete-WAM~\cite{yao2026discrete}
            & \multicolumn{1}{c|}{SV}
            & {98.5}
            & {98.2}
            & {99.7}
            & {99.8}
            & {90.5}
            & {97.9}
            & {97.2}
            & {98.3}
            & {78.1}
            & {90.4} \\
			
            DriveWAM~\cite{shi2026drivewam}
            & \multicolumn{1}{c|}{SV}
            & {98.4}
            & {96.7}
            & {99.7}
            & {99.8}
            & {97.7}
            & {97.7}
            & {97.7}
            & {98.3}
            & {87.9}
            & {92.2} \\

            ByondDrive~\cite{wang2026beyond}
            & {SV}
            & {98.2}
            & {97.7}
            & {99.4}
            & {99.8}
            & {97.9}
            & {97.8}
            & {97.7}
            & {98.3}
            & {87.6}
            & {92.9} \\

            \rowcolor[HTML]{FEF0C7}
            \textbf{Ours}
            & \multicolumn{1}{c|}{SV}
            & {98.1}
            & {98.4}
            & {99.5}
            & {99.8}
            & {97.8}
            & {97.5}
            & {97.3}
            & {98.3}
            & {87.6}
            & {\textbf{93.3}} \\
            
			\bottomrule
	\end{tabular}}
\end{table*}

\subsection{Additional Qualitative Results}
\label{sec:supp_results_add}

\subsubsection{Junction Interaction in HUGSIM}
\label{sec:supp_hugsim_visual}

\begin{figure*}[!t]
    \centering
    \includegraphics[width=\textwidth]{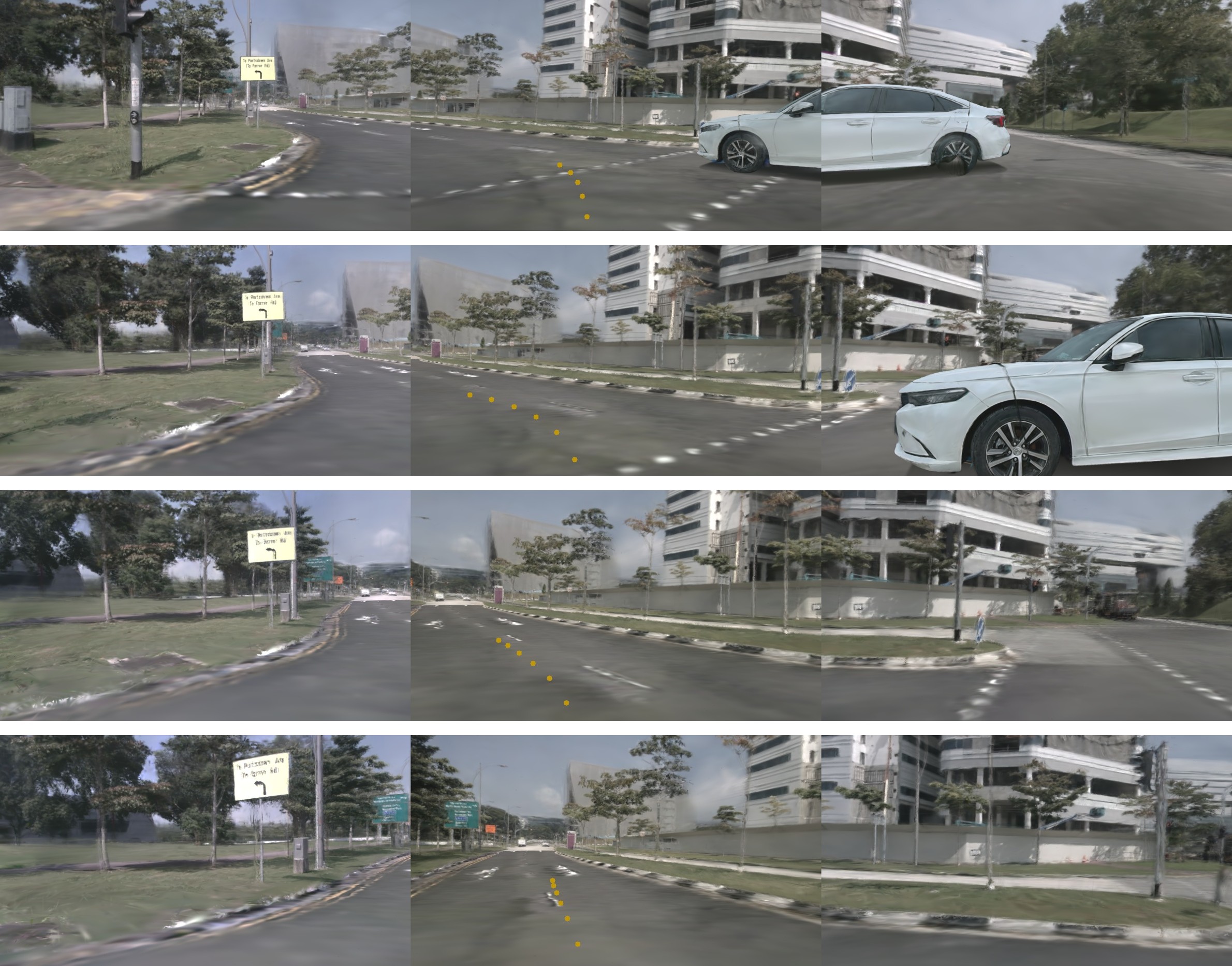}
    \caption{Additional closed-loop rollout near a junction in HUGSIM. The four rows are ordered chronologically from top to bottom; each row contains three rendered camera views. Yellow dots show the overlaid trajectory in the center view. A nearby white vehicle is visible in the early snapshots, followed by the ego vehicle's continued motion along the road.}
    \label{fig:supp_hugsim_junction}
\end{figure*}

Figure~\ref{fig:supp_hugsim_junction} shows an additional closed-loop interaction in HUGSIM~\cite{zhou2025hugsim}. A white vehicle appears near the junction in the first two snapshots and moves out of the displayed views as the ego vehicle advances along the road. The three rendered views at each time point show the surrounding traffic and road geometry, while the yellow trajectory overlay changes with the observations. This sequence complements the oncoming-vehicle example in the main paper by showing another traffic interaction during closed-loop execution.

\subsubsection{Additional Trajectory Comparisons}
\begin{figure*}[!ht]
\centering
\includegraphics[width=\textwidth]
{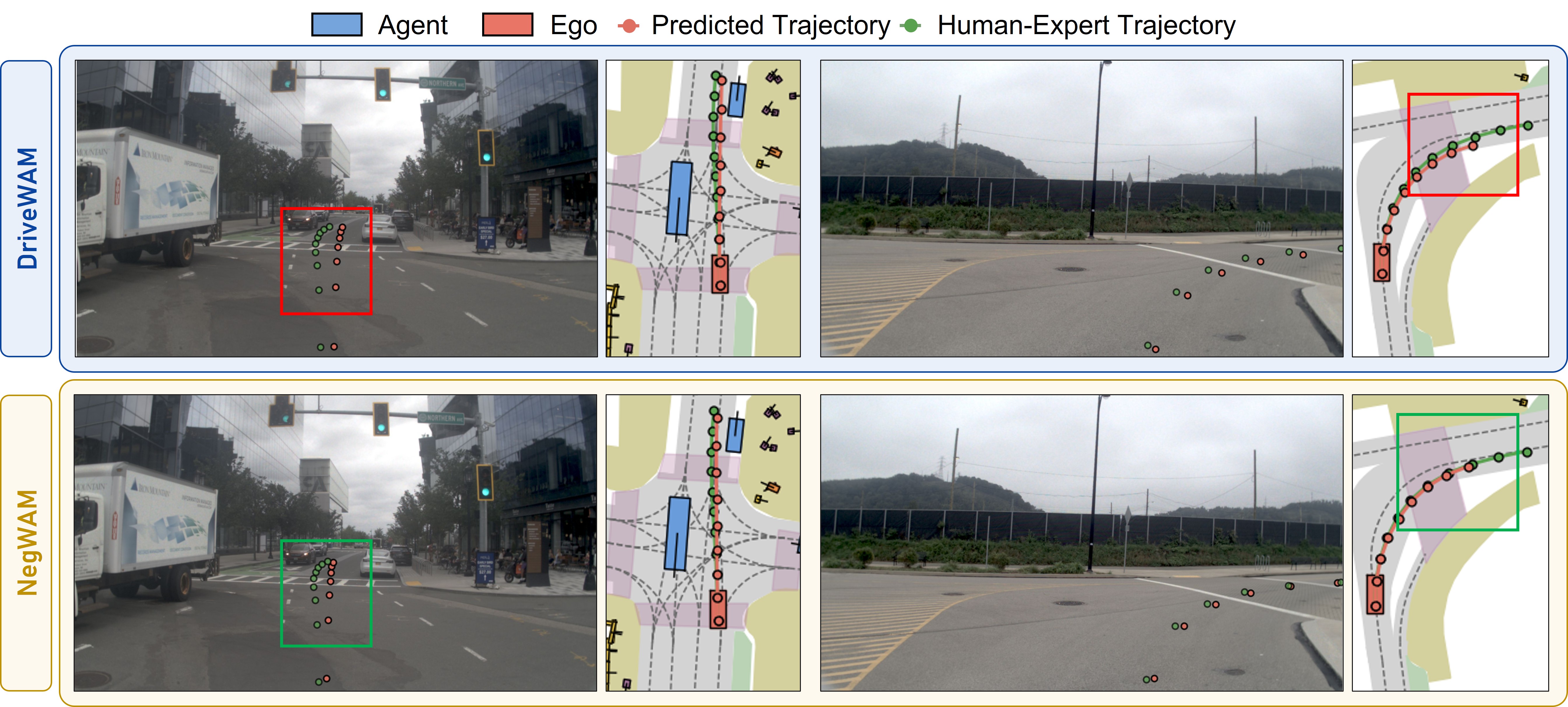}
\caption{
Additional successful cases.
Comparison between DriveWAM~\cite{shi2026drivewam} (top) and NegWAM (bottom) in an
intersection-straight scenario with a stopped roadside vehicle and an
intersection left-turn scenario. NegWAM safely avoids the stopped
vehicle and maintains its trajectory within the drivable area, while
the baseline produces collision-prone or off-road trajectories.}
\label{fig:fig_s1_qualita}
\end{figure*}

Figure~\ref{fig:fig_s1_qualita} compares DriveWAM with NegWAM in two scenarios beyond those shown in the main paper. In the intersection-straight case, NegWAM's trajectory bypasses a stopped roadside vehicle, while DriveWAM predicts a path close to the obstacle. In the left-turn case, NegWAM's trajectory remains within the drivable surface, whereas the baseline trajectory leaves it. The two examples illustrate differences in obstacle clearance and adherence to road geometry.

\subsubsection{Failure Cases}

\begin{figure*}[!ht]
\centering
\includegraphics[width=\textwidth]
{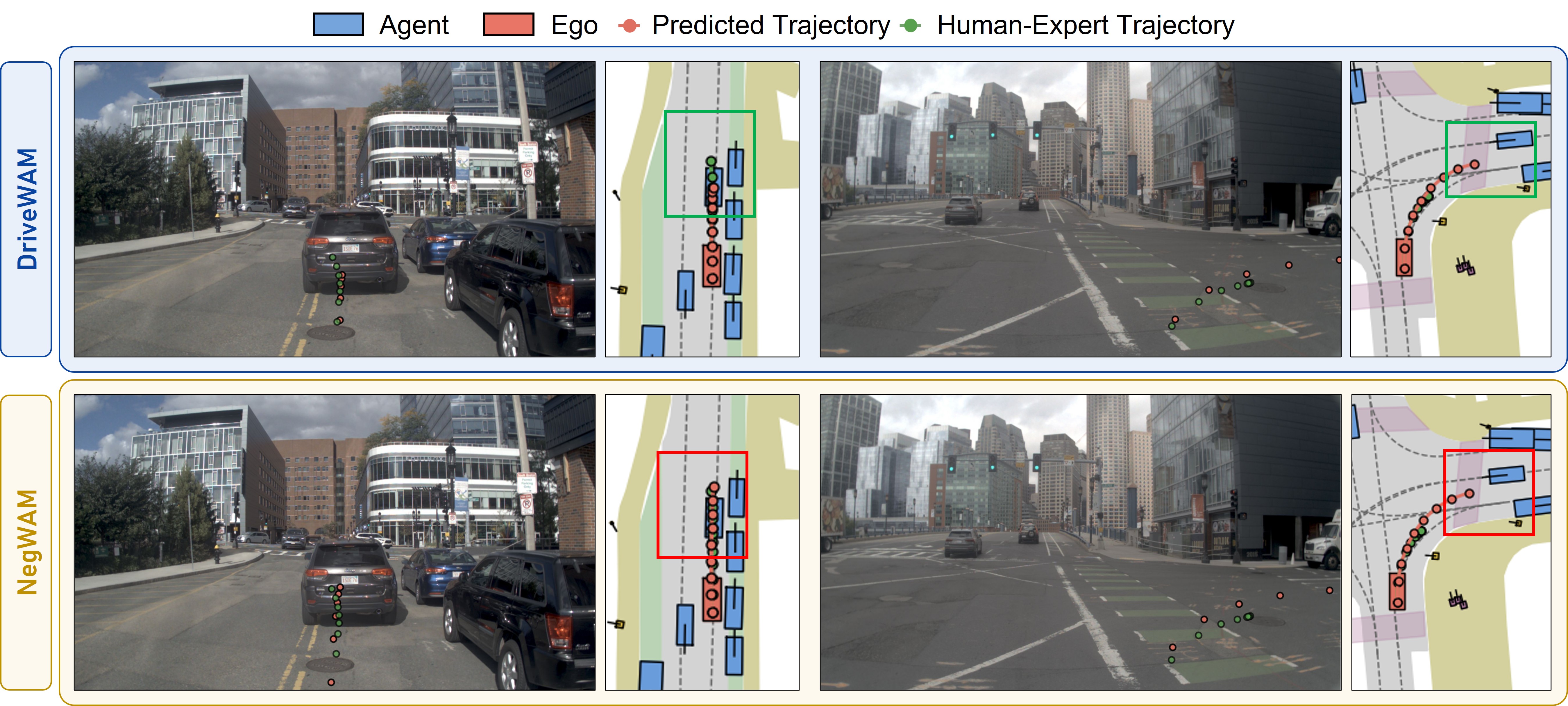}
\caption{
Representative failure cases.
Comparison between DriveWAM (top) and NegWAM (bottom). In the first case, NegWAM fails to follow the expert's leftward overtaking maneuver and collides with the stopped leading vehicle, whereas DriveWAM avoids the collision through conservative deceleration. In the second case, NegWAM maintains insufficient clearance from a vehicle in the target lane while turning, resulting in a collision.}
\label{fig:fig_s2_failure}
\end{figure*}

Figure~\ref{fig:fig_s2_failure} documents two cases in which NegWAM performs worse than DriveWAM. In the first, the expert overtakes a stopped leading vehicle on the left, but NegWAM continues straight and collides with it. DriveWAM slows behind the vehicle and avoids the collision. In the second, NegWAM turns with insufficient clearance from a vehicle in the target lane, while DriveWAM maintains a wider margin. These failures identify limitations in maneuver selection and interaction clearance that remain after post-training.

\subsection{Scope of the Negative Supervision}
\label{sec:supp_limit}

\paragraph{Hard-Negative Coverage.}
The learned safety boundary depends on the near-expert failures covered by the fixed trajectory vocabulary. Rare maneuvers and interactions outside that candidate space receive no direct negative supervision. Vocabulary-based retrieval provides broad trajectory coverage but limited adaptation to individual scenes~\cite{li2025gtrs}; diversity-aware failure generation is a complementary direction~\cite{wang2026beyond}.

\paragraph{Evaluator and Geometric-Distance Dependence.}
The zero-EPDMS criterion assigns a common failure label to different violation types and severities. Both hard-negative objectives therefore inherit the offline evaluator's assumptions. ADE additionally measures geometric proximity rather than interaction risk: similar paths can have different outcomes depending on timing, relative velocity, and road geometry~\cite{dauner2024navsim}. The rider branch uses scene context, but its negative labels still come from this same construction process.

\paragraph{Broader Impact.}
The additional failure cases show that benchmark gains can coexist with unsafe individual decisions. The evidence remains benchmark-based and provides no formal safety guarantee. Deployment requires independent closed-loop validation, evaluation across traffic conditions and road-user categories, and safeguards for hazardous or unfamiliar situations.

\subsection{License of Assets}
\label{sec:supp_license}

All training and evaluation data are obtained from publicly released datasets, including nuPlan~\cite{karnchanachari2024nuplan}, OpenScene~\cite{openscene2023}, and NAVSIM v2~\cite{cao2025navsimv2}, and are used in accordance with their respective licenses and terms of use. Our implementation further builds upon publicly available model assets from LingBot-VA~\cite{li2026causal}, Wan2.2-TI2V-5B~\cite{wan2025wan}, and DriveWAM~\cite{shi2026drivewam}. All third-party assets are used solely for non-commercial academic research and in compliance with the licenses accompanying their original releases.

\end{document}